\documentclass[11pt]{article}

\usepackage{fullpage}
\usepackage[ruled, linesnumbered, vlined, commentsnumbered]{algorithm2e}
\usepackage{graphicx,subfig} 
\usepackage{amsfonts,amssymb,amsmath,amsthm,amsopn}	
\usepackage{hhline,booktabs,colortbl,multirow,tabularx,threeparttable} 

\usepackage{enumerate}
\usepackage{authblk}
\usepackage{footnote}
\usepackage{hyperref}
\usepackage{prettyref}
\usepackage{cite}
\usepackage{setspace}
\usepackage{color}
\usepackage{xcolor}  
\usepackage{scrpage2}
\usepackage{geometry}

\usepackage{tikz}
\usepackage{pgfplots}
\usetikzlibrary{positioning,shapes,shadows,arrows,calc}
\tikzstyle{component}=[rectangle, draw=black, rounded corners, fill=blue!40, drop shadow, text centered, anchor=north, text=white, minimum height=1cm]
\tikzstyle{arrow}=[->, thick]

\pgfplotsset{compat=1.12}
\usetikzlibrary{intersections}
\usetikzlibrary{pgfplots.statistics}
\usepgfplotslibrary{fillbetween}

\definecolor{myblue}{RGB}{34,31,217}
\definecolor{mycyan}{gray}{.7}
\definecolor{Gray}{gray}{0.9}

\newcommand{\pref}{\prettyref}

\newrefformat{fig}{Fig.~\ref{#1}}
\newrefformat{tab}{Table~\ref{#1}}
\newrefformat{sec}{Section~\ref{#1}}
\newrefformat{app}{Appendix~\ref{#1}}
\newrefformat{alg}{Algorithm~\ref{#1}}
\newrefformat{property}{Property~\ref{#1}}
\newrefformat{theorem}{Theorem~\ref{#1}}
\newrefformat{corollary}{Corollary~\ref{#1}}
\newrefformat{proposition}{Proposition~\ref{#1}}
\newrefformat{def}{Definition~\ref{#1}}
\newrefformat{eq}{equation~(\ref{#1})}

\usepackage{lscape}

\begin{document}

\title{\vspace{-1ex}\LARGE\textbf{Which Surrogate Works for Empirical Performance Modelling? A Case Study with Differential Evolution}\footnote{This manuscript is submitted for possible publication. The reviewer can use this manuscript in peer review.}}

\author[1]{\normalsize Ke Li}
\author[2]{\normalsize Zilin Xiang}
\author[3]{\normalsize Kay Chen Tan}
\affil[1]{\normalsize Department of Computer Science, University of Exeter}
\affil[2]{\normalsize School of Computer Science and Engineering\\ University of Electronic Science and Technology of China}
\affil[3]{\normalsize Department of Computer Science, City University of Hong Kong}
\affil[$\ast$]{\normalsize Email: \texttt{k.li@exeter.ac.uk, zilin.xiang@hotmail.com, kaytan@cityu.edu.hk}}

\date{}
\maketitle

\vspace{-3ex}
{\normalsize\textbf{Abstract: } }It is not uncommon that meta-heuristic algorithms contain some intrinsic parameters, the optimal configuration of which is crucial for achieving their peak performance. However, evaluating the effectiveness of a configuration is expensive, as it involves many costly runs of the target algorithm. Perhaps surprisingly, it is possible to build a cheap-to-evaluate surrogate that models the algorithm's empirical performance as a function of its parameters. Such surrogates constitute an important building block for understanding algorithm performance, algorithm portfolio/selection, and the automatic algorithm configuration. In principle, many off-the-shelf machine learning techniques can be used to build surrogates. In this paper, we take the differential evolution (DE) as the baseline algorithm for proof-of-concept study. Regression models are trained to model the DE's empirical performance given a parameter configuration. In particular, we evaluate and compare four popular regression algorithms both in terms of how well they predict the empirical performance with respect to a particular parameter configuration, and also how well they approximate the parameter \textit{versus} the empirical performance landscapes.

{\normalsize\textbf{Keywords: } }Empirical performance modelling, parameter configuration, landscape analysis, differential evolution.


\section{Introduction}
\label{sec:introduction}

Meta-heuristic algorithms are normally accompanied by some parameters which can influence their search behaviour on various optimisation problems. Parameter optimisation (PO) aims to find a best possible parameter configuration $\mathbf{\theta}^{\ast}$ from the parameter space $\Theta$, which consists of all possible configurations, of the target algorithm and helps it achieve its peak performance on a black-box optimisation problem. Formally, given an algorithm, PO can be defined as the following black-box meta-optimisation problem:
\begin{equation}
\begin{array}{l l}
\mathrm{minimize} \quad \mathcal{L}(f(\mathbf{x}),\theta)\\
\mathrm{subject\ to} \quad \theta\in\Theta
\end{array}
\label{eq:loss}
\end{equation}
where $f(\mathbf{x})$ is the optimisation problem under consideration, and $\mathbf{x}\in\mathbb{R}^d$ is a decision variable. $\mathcal{L}(f(\mathbf{x}),\theta)$ is the performance measure associated with a configuration $\theta$ of the target algorithm. In particular, it can either be the runtime cost (e.g. the CPU wall time and/or the number of function evaluations) or the error of the solution found by the target algorithm.

PO is a challenging black-box meta-optimisation problem. First, its landscape is complex and change with the target algorithm when solving different problems. Second, the parameters associated with the target algorithm can have various types (e.g. numerical, integer and categorical) and the number of parameters can be potentially large depending on the algorithm specification. In addition, PO is intrinsically expensive as it requires to explore $\Theta$ by running the target algorithm with different configurations, where evaluating the effectiveness of a configuration will in turn cost a large amount of function evaluations and/or CPU wall time. In the evolutionary computation (EC) community, constructing a cheap-to-evaluate surrogate in lieu of calling the physically expensive objective function has been widely accepted as an effective way for expensive optimisation~\cite{Jin11}. The design and analysis of computer experiments in statistics also uses surrogate models to either fit a global model of the overall landscape or sequentially identify the global optimum of the underlying function~\cite{SantnerWN03}. In the automatic parameter configuration field, sequential model-based Bayesian optimisation methods~\cite{Bartz-BeielsteinLP05,HutterHL11,ThorntonHHL13} have shown strong performance in PO, compared to some traditional methods like grid search and random search~\cite{BergstraB12} and can compete or even surpass the results tuned by experienced human experts. Moreover, regression models have been extensively used in meta-learning to predict the algorithm performance across various datasets~\cite{Reif14}. It is worth to note that all these lines of research need to construct surrogate models of a computationally expensive and complex function in order to inform an active learning criterion that identifies new inputs to evaluate.

The problem of PO has a long history dating back to the 90s~\cite{KohaviJ95}. Recently, it becomes increasingly popular in both meta-heuristics (e.g.~\cite{Bartz-BeielsteinLP05,HutterHL11,BlotHJKT16,LopDubPerStuBir2016irace,LiFKZ14}) and machine learning (e.g.~\cite{SnoekLA12,ThorntonHHL13,SandersG17,CaoKWL12,LiWKC13,CaoKWL14,CaoKWLLK15}) communities, especially with the development of emerging automated machine learning~\cite{automl}. In this paper, instead of developing new algorithms for PO, we focus on studying surrogate models, which sit in the core of the model-based PO framework. We take the differential evolution (DE)~\cite{StornP97,LiKWCR12}, one of the most popular black-box optimiser in the EC community, as the baseline algorithm. To obtain the empirical performance data on a given optimisation problem, we evaluate the performance of DE with respect to 5,940 parameter configurations in an expensive offline phase. The collected performance data are used to train a regression model and to validate its generalisation ability for predicting empirical performance of unseen parameter configurations. Here we consider four off-the-shelf regression algorithms for empirical performance modelling. In particular, we evaluate and compare their abilities in terms of how well they predict the empirical performance with respect to a particular parameter configuration, and also how well they approximate the parameter configuration \textit{versus} the empirical performance landscapes. We envisage that this aspect will shed light on the study of the characteristics of surrogate models in future.

The rest of this paper is organised as follows. \pref{sec:method} describes the methodologies that we used to setup the experiments. \pref{sec:experiments} presents and analyses the experimental results. Finally, \pref{sec:conclusion} concludes this paper and provides some future directions.


\section{Methodology}
\label{sec:method}

This section mainly describes the benchmark problems chosen in our empirical studies, the baseline algorithm DE and its corresponding parameters, the performance measure used to evaluate the quality of a particular parameter configuration, the method used to collect the algorithm performance data, and the regression algorithms used to build surrogates for modelling the empirical performance.

\subsection{Benchmark Problems}
\label{sec:benchmark}

In this paper, we consider choosing six widely used elementary test problems (i.e. sphere, ellipsoid, rosenbrock, ackley, griewank and rastrigin) and the first fourteen test problems (i.e. excluding those hybrid composite functions) from the CEC 2005 competition~\cite{SuganthanHDLCAT05} to constitute the benchmark problems. To facilitate the notation in~\pref{sec:experiments}, the six elementary functions are denoted as F1 to F6 and those from the CEC 2005 competition are denoted as F7 to F20. Note that these test problems have various characteristics. In particular, F1, F2 and F7 to F11 are unimodal functions while the others are multi-modal functions. All test problems have analytically defined continuous objective functions with a known global optimum. The number of variables of each test problem varies from 2 to 30 (in particular $d\in\{2,10,30\}$) and the range of variables is set according to their original paper.

\subsection{DE and its Parameters}
\label{sec:de}

DE~\cite{StornP97} is one of the most popular black-box optimisation algorithm in the EC community including evolutionary multi-objective optimisation~\cite{LiZKLW14,LiKD15,LiKZD15,LiDZK15,LiDZZ17,ChenLY18,LiCMY18}. One of the major reasons that contributes to its success is its simple structure. For a vanilla DE, an offspring solution $\mathbf{x}^c$ is generated by a two-step procedure. First, a trial vector $\overline{\mathbf{x}}$ is generated as:
\begin{equation}
\overline{\mathbf{x}}=\mathbf{x}^1+F\times(\mathbf{x}^2-\mathbf{x}^3)
\label{eq:trial}
\end{equation}
where $F\in(0,3]$, known as the evolution step size, is a parameter of DE. $\mathbf{x}^1$, $\mathbf{x}^2$ and $\mathbf{x}^3$ are randomly chosen from the parent population. Afterwards, $\mathbf{x}^c$ is generated as:
\begin{equation}
x^c_i=
\begin{cases}
\overline{x_i} & \quad \text{if } (rand<CR)\lor(i=j)\\
x_i   & \quad \text{otherwise}
\end{cases}
\end{equation}
where $i\in\{1,\cdots, d\}$, $j$ is an integer randomly chosen from 1 to $d$. $\mathbf{x}$ is the parent solution under consideration. $rand$ is a random number chosen from 0 to 1, and $CR\in[0,1]$, known as the crossover rate, is another parameter of DE. In addition, the population size $NP\in\mathbb{N}$ is also a parameter.

Many studies have demonstrated that the performance of DE is highly sensitive to its parameter settings~\cite{DasS11}. During the past decade, many efforts have been devoted to the development of advanced DE variants that are able to adaptively set the parameters on the fly~\cite{BrestGBMZ06,QinHS09,LiFK11} and/or find a good configuration in an offline manner~\cite{BelkhirDSS16}. Since the major purpose of this paper is to investigate the ability of building the surrogate for modelling the empirical performance of an algorithm with respect to its corresponding parameter configurations, we focus on the vanilla DE~\cite{StornP97} which is simple yet without losing the generality of the observations. Obviously, $NP$ is an integer parameter, while $F$ and $CR$ are numerical parameters.


\subsection{Performance Measure}
\label{sec:measure}

As the global optimum of each test problem is known a priori, this paper uses the approximation error to evaluate the empirical performance of a particular parameter configuration. Specifically, it is computed as:
\begin{equation}
    \Psi(f(\mathbf{x}),\theta)=f(\mathbf{x})-f(\mathbf{x}^{\ast})
\end{equation}
where $\theta$ is a parameter configuration of DE, $\mathbf{x}$ is the best-so-far solution found by the DE with the parameter configuration $\theta$, and $\mathbf{x}^\ast$ is the global optimum. Since DE is a stochastic algorithm, each parameter configuration needs to be repeated more than one time in practice. Thus, the performance of a parameter configuration $\theta$ is measured as an averaged approximation error:
\begin{equation}
	\mathcal{L}(f(\mathbf{x}),\theta)=\frac{1}{n}\sum_{i=1}^n\Psi_i(f(\mathbf{x}),\theta)
\end{equation}
where $\Psi_i(f(\mathbf{x}),\theta)$ is the approximation error of a configuration $\theta$ at the $i$-th run and $n$ is the number of repetitions of experiments with $\theta$ where we set $n=31$ in our experiments.

\subsection{Data Collection}
\label{sec:data}

In principle, algorithm performance data used to construct the surrogate model of an algorithm's empirical performance can be obtained by any means. Since this paper aims to investigate the overall surrogate modelling ability of an algorithm's performance with respect to its parameter space, we are interested in every corner of the space. To this end, the parameter space is sampled in a grid manner, where we chose 9 different $NP$ settings, i.e. $NP=i\times d$, $i\in\{2,\cdots,10\}$, 60 different values for $F\in(0,3]$ with a step size 0.05, and 11 different values for $CR\in[0,1]$ with a step size 0.1. Therefore, there are 5,940 different parameter configurations in total.

\subsection{Regression Algorithms for Surrogate Modelling}
\label{sec:surrogate}

In this paper, four regression algorithms, i.e. Gaussian process (GP), random forest (RF), support vector machine for regression (SVR), radial basis function networks (RBFN), are considered as the candidates for surrogate modelling of DE's empirical performance. Note that these regression algorithms have been widely used in the model-based PO in the algorithm configuration literature~\cite{HutterXHL14,WuKJLZ17,WuLKZZ18}.

To construct a surrogate model on a particular problem instance, each of these four models is trained on the performance data (only 70\% of them are used for training while the remaining 30\% are used for testing) collected by running the DE algorithm with various parameter configurations on each problem instance as introduced in~\pref{sec:data}. Note that learning a surrogate model is no free lunch, as each regression algorithm also requires some hyper-parameters to be tuned. To identify the best possible configurations for each regression algorithm, we apply the random search~\cite{BergstraB12} to explore the hyper-parameter space. Specifically, as for GP, we need to choose an appropriate kernel among RBF, rational quadratic and Mat\'ern; as for RF, the number of trees in a forest is chosen from 2 to 100, the minimum number of samples required to split an internal node is chosen from 2 to 11, the number of features to consider when looking for the best split is set in the range $[0.001,1]$, the criterion used to measure the quality of a split is either mean squared error or mean absolute error and the minimum number of samples required to be at a leaf node is chosen from 1 to 11; as for SVR, the kernel is chosen between RBF and Sigmoid, the maximal margin $\epsilon$ is chosen from $[0.01,1]$, the regularisation parameter $C$ is set in between 1 and 10, and $\gamma$ is chosen from $[0.01,1]$ if RBF is used as the kernel. A 5-fold cross-validation (using 80\% of the training data for training and the remaining 20\% data for testing) is used to evaluate the training performance of a particular hyper-parameter configuration of a regression algorithm. To have a fair comparison, all surrogate modelling procedures are implemented by \texttt{scikit-learn}, a machine learning toolbox in Python\footnote{\url{https://scikit-learn.org/stable/}}.


\section{Experiments and Results}
\label{sec:experiments}

In this section, we will present and compare experimental evaluations of the quality of surrogates constructed by different regression algorithms introduced in~\pref{sec:surrogate}. The experimental results are analysed according to the following three research questions (RQs).

\begin{description}
    \item[\underline{\textbf{RQ1:}}] \textbf{\textit{Which surrogate model works best for empirical performance modelling on various kinds of benchmark problems?}}
    \item[\underline{\textbf{RQ2:}}] \textbf{\textit{Does the empirical performance predicted by a surrogate model follow the order as the ground truth?}}
    \item[\underline{\textbf{RQ3:}}] \textbf{\textit{How does the empirical performance landscape fit by a surrogate model compare with the ground truth?}}
\end{description}

\subsection{Comparisons of Different Surrogate Models}
\label{sec:RMSE_comparison}

Bearing the RQ1 in mind, this section empirically compares the generalisation performance of four regression algorithms on unseen parameter configurations. In particular, the root mean square error (RMSE) is used to measure the generalisation performance and it is calculated as:
\begin{equation}
    RMSE=\sqrt{\frac{\sum_{i=1}^{\hat{n}}(\hat{\mathcal{L}}(f(\mathbf{x}),\theta_i)-\mathcal{L}(f(\mathbf{x}),\theta_i))^2}{\hat{n}}}
    \label{eq:rmse}
\end{equation}
where $\hat{\mathcal{L}}(f(\mathbf{x}),\theta_i)$ is the approximation error of a parameter configuration $\theta_i$ estimated by a surrogate model; while $\mathcal{L}(\mathbf{x},\theta_i)$ is the observed approximation error of $\theta_i$, $i\in\{1,\cdots,\hat{n}\}$ and $\hat{n}$ is the number of data in the testing set.

From the results shown in Tables~\ref{tab:metric_2D} to \ref{tab:metric_30D}, we clearly see that GP and RF are the best regression algorithms to build the surrogate for modelling the empirical performance. RBFN is slightly worse than GP and RF, while SVR is the worst choice except on F14 when $d=2$. Note that our observations of promising performance of GP and RF are also in line with some results reported in the contemporary algorithm configuration literature~\cite{HutterXHL14}. Furthermore, we find that the performance of different regression algorithms are consistent across different dimensions. This makes sense as a surrogate model is built upon the parameter configurations themselves, which are independent from the problem instances. In addition, we find that the RMSE dramatically increases with the dimensionality of the underlying problem. This can be explained as the significant degeneration of the performance of DE with the dimensionality which in term largely increases the approximation errors.

\begin{table*}[htbp]
    \centering
    \caption{Comparisons of RMSE, PCC and SRCC obtained by four regression algorithms on benchmark problems $(d=2)$}
    \resizebox{\columnwidth}{!}{
    \begin{tabular}{c|c|c|c|c|c||c|c|c|c|c|c}
    \hline
    \textbf{Problem} & \textbf{Metric} & \textbf{GP} & \textbf{RBFN} & \textbf{RF} & \textbf{SVR} & \textbf{Problem} & \textbf{Metric} & \textbf{GP} & \textbf{RBFN} & \textbf{RF} & \textbf{SVR} \\
    \hline
    \multirow{1}[6]{*}{F1} & RMSE  & \cellcolor[rgb]{ .651,  .651,  .651}\textbf{1.3605E-1} & 1.3692E-1 & 1.5620E-1 & 6.4804E-1 & \multirow{1}[6]{*}{F11} & RMSE  & \cellcolor[rgb]{ .651,  .651,  .651}\textbf{1.1228E+1} & 1.2321E+1 & 1.1611E+1 & 3.0824E+1 \\
\cline{2-6}\cline{8-12}          & PCC   & \cellcolor[rgb]{ .651,  .651,  .651}\textbf{9.9025E-1} & 9.8920E-1 & 9.8818E-1 & 7.1022E-1 &       & PCC   & \cellcolor[rgb]{ .651,  .651,  .651}\textbf{9.8613E-1} & 9.8325E-1 & 9.8563E-1 & 9.1249E-1 \\
\cline{2-6}\cline{8-12}          & SRCC  & 9.0219E-1 & 8.4648E-1 & \cellcolor[rgb]{ .651,  .651,  .651}\textbf{9.6439E-1} & 7.2813E-1 &       & SRCC  & 8.1685E-1 & 8.0941E-1 & \cellcolor[rgb]{ .651,  .651,  .651}\textbf{8.6094E-1} & 8.1033E-1 \\
    \hline
    \multirow{1}[6]{*}{F2} & RMSE  & 5.6003E+0 & 6.8769E+0 & \cellcolor[rgb]{ .651,  .651,  .651}\textbf{4.8678E+0} & 1.0089E+1 & \multirow{1}[6]{*}{F12} & RMSE  & \cellcolor[rgb]{ .651,  .651,  .651}\textbf{1.8104E+7} & 1.9118E+7 & 2.0662E+7 & 1.1799E+8 \\
\cline{2-6}\cline{8-12}          & PCC   & 9.7771E-1 & 9.6623E-1 & \cellcolor[rgb]{ .651,  .651,  .651}\textbf{9.8320E-1} & 9.3107E-1 &       & PCC   & \cellcolor[rgb]{ .651,  .651,  .651}\textbf{9.8769E-1} & 9.8643E-1 & 9.8516E-1 & 3.0080E-1 \\
\cline{2-6}\cline{8-12}          & SRCC  & 8.4210E-1 & 7.4855E-1 & \cellcolor[rgb]{ .651,  .651,  .651}\textbf{9.4101E-1} & 8.5432E-1 &       & SRCC  & 6.0502E-1 & 4.0852E-1 & \cellcolor[rgb]{ .651,  .651,  .651}\textbf{8.6027E-1} & 7.2927E-1 \\
    \hline
    \multirow{1}[6]{*}{F3} & RMSE  & 4.6287E+2 & \cellcolor[rgb]{ .651,  .651,  .651}\textbf{4.5074E+2} & 4.8185E+2 & 6.4441E+2 & \multirow{1}[6]{*}{F13} & RMSE  & \cellcolor[rgb]{ .651,  .651,  .651}\textbf{1.4518E+0} & 1.9035E+0 & 2.9745E+0 & 1.1412E+1 \\
\cline{2-6}\cline{8-12}          & PCC   & 7.4558E-1 & \cellcolor[rgb]{ .651,  .651,  .651}\textbf{7.6157E-1} & 7.5379E-1 & 3.9173E-1 &       & PCC   & \cellcolor[rgb]{ .651,  .651,  .651}\textbf{9.9909E-1} & 9.9843E-1 & 9.9647E-1 & 9.5266E-1 \\
\cline{2-6}\cline{8-12}          & SRCC  & 8.6529E-1 & 7.2109E-1 & \cellcolor[rgb]{ .651,  .651,  .651}\textbf{9.7141E-1} & 9.4045E-1 &       & SRCC  & \cellcolor[rgb]{ .651,  .651,  .651}\textbf{8.9138E-1} & 7.7192E-1 & 8.8296E-1 & 6.3004E-1 \\
    \hline
    \multirow{1}[6]{*}{F4} & RMSE  & 5.6182E-1 & 9.3965E-1 & \cellcolor[rgb]{ .651,  .651,  .651}\textbf{5.2925E-1} & 9.7513E-1 & \multirow{1}[6]{*}{F14} & RMSE  & 1.0581E+0 & 1.3978E+0 & 1.0491E+0 & \cellcolor[rgb]{ .651,  .651,  .651}\textbf{1.0448E+0} \\
\cline{2-6}\cline{8-12}          & PCC   & 9.8979E-1 & 9.7194E-1 & \cellcolor[rgb]{ .651,  .651,  .651}\textbf{9.9103E-1} & 9.7017E-1 &       & PCC   & 9.3824E-1 & 8.9687E-1 & 9.3839E-1 & \cellcolor[rgb]{ .651,  .651,  .651}\textbf{9.3922E-1} \\
\cline{2-6}\cline{8-12}          & SRCC  & 9.7495E-1 & 9.6736E-1 & \cellcolor[rgb]{ .651,  .651,  .651}\textbf{9.8303E-1} & 9.5503E-1 &       & SRCC  & \cellcolor[rgb]{ .651,  .651,  .651}\textbf{9.3023E-1} & 9.0572E-1 & 9.2777E-1 & 9.2696E-1 \\
    \hline
    \multirow{1}[6]{*}{F5} & RMSE  & 1.4404E-2 & 1.9011E-2 & \cellcolor[rgb]{ .651,  .651,  .651}\textbf{1.3800E-2} & 1.5263E-2 & \multirow{1}[6]{*}{F15} & RMSE  & \cellcolor[rgb]{ .651,  .651,  .651}\textbf{5.3994E-1} & 6.1380E-1 & 5.6273E-1 & 7.0871E-1 \\
\cline{2-6}\cline{8-12}          & PCC   & 9.4829E-1 & 9.1330E-1 & \cellcolor[rgb]{ .651,  .651,  .651}\textbf{9.5213E-1} & 9.4119E-1 &       & PCC   & \cellcolor[rgb]{ .651,  .651,  .651}\textbf{9.8694E-1} & 9.8310E-1 & 9.8595E-1 & 9.7769E-1 \\
\cline{2-6}\cline{8-12}          & SRCC  & 9.5309E-1 & 9.3783E-1 & \cellcolor[rgb]{ .651,  .651,  .651}\textbf{9.5406E-1} & 9.4455E-1 &       & SRCC  & 9.8333E-1 & 9.8193E-1 & \cellcolor[rgb]{ .651,  .651,  .651}\textbf{9.8491E-1} & 9.7681E-1 \\
    \hline
    \multirow{1}[6]{*}{F6} & RMSE  & 5.1460E-1 & 5.8459E-1 & \cellcolor[rgb]{ .651,  .651,  .651}\textbf{4.8990E-1} & 8.5190E-1 & \multirow{1}[6]{*}{F16} & RMSE  & 8.2311E-1 & 9.4311E-1 & \cellcolor[rgb]{ .651,  .651,  .651}\textbf{7.6776E-1} & 1.0465E+0 \\
\cline{2-6}\cline{8-12}          & PCC   & 9.8631E-1 & 9.8242E-1 & \cellcolor[rgb]{ .651,  .651,  .651}\textbf{9.8764E-1} & 9.6360E-1 &       & PCC   & 9.8415E-1 & 9.7927E-1 & \cellcolor[rgb]{ .651,  .651,  .651}\textbf{9.8628E-1} & 9.7440E-1 \\
\cline{2-6}\cline{8-12}          & SRCC  & 9.8310E-1 & 9.8082E-1 & \cellcolor[rgb]{ .651,  .651,  .651}\textbf{9.8486E-1} & 9.7742E-1 &       & SRCC  & 9.8511E-1 & 9.8031E-1 & \cellcolor[rgb]{ .651,  .651,  .651}\textbf{9.8572E-1} & 9.7966E-1 \\
    \hline
    \multirow{1}[6]{*}{F7} & RMSE  & \cellcolor[rgb]{ .651,  .651,  .651}\textbf{5.3570E+1} & 5.7449E+1 & 6.3230E+1 & 2.8793E+2 & \multirow{1}[6]{*}{F17} & RMSE  & 9.0189E-2 & 1.2394E-1 & \cellcolor[rgb]{ .651,  .651,  .651}\textbf{7.6712E-2} & 9.3559E-2 \\
\cline{2-6}\cline{8-12}          & PCC   & \cellcolor[rgb]{ .651,  .651,  .651}\textbf{9.9132E-1} & 9.9003E-1 & 9.8804E-1 & 7.5629E-1 &       & PCC   & 9.8626E-1 & 9.7430E-1 & \cellcolor[rgb]{ .651,  .651,  .651}\textbf{9.9011E-1} & 9.8523E-1 \\
\cline{2-6}\cline{8-12}          & SRCC  & 9.2604E-1 & 9.1303E-1 & \cellcolor[rgb]{ .651,  .651,  .651}\textbf{9.5654E-1} & 8.9479E-1 &       & SRCC  & 9.8428E-1 & 9.7754E-1 & \cellcolor[rgb]{ .651,  .651,  .651}\textbf{9.8806E-1} & 9.8360E-1 \\
    \hline
    \multirow{1}[6]{*}{F8} & RMSE  & \cellcolor[rgb]{ .651,  .651,  .651}\textbf{5.3072E+1} & 6.1123E+1 & 6.5377E+1 & 2.9297E+2 & \multirow{1}[6]{*}{F18} & RMSE  & 1.0509E+2 & 1.0808E+2 & \cellcolor[rgb]{ .651,  .651,  .651}\textbf{9.5303E+1} & 2.5797E+2 \\
\cline{2-6}\cline{8-12}          & PCC   & \cellcolor[rgb]{ .651,  .651,  .651}\textbf{9.9178E-1} & 9.8911E-1 & 9.8790E-1 & 7.5211E-1 &       & PCC   & 9.6875E-1 & 9.6675E-1 & \cellcolor[rgb]{ .651,  .651,  .651}\textbf{9.7438E-1} & 8.0866E-1 \\
\cline{2-6}\cline{8-12}          & SRCC  & 9.6614E-1 & 9.5066E-1 & \cellcolor[rgb]{ .651,  .651,  .651}\textbf{9.6749E-1} & 9.0941E-1 &       & SRCC  & 9.6516E-1 & 9.5743E-1 & \cellcolor[rgb]{ .651,  .651,  .651}\textbf{9.7273E-1} & 9.2020E-1 \\
    \hline
    \multirow{1}[6]{*}{F9} & RMSE  & \cellcolor[rgb]{ .651,  .651,  .651}\textbf{1.0347E+7} & 1.4066E+7 & 1.1402E+7 & 6.3483E+7 & \multirow{1}[6]{*}{F19} & RMSE  & 4.5497E+0 & \cellcolor[rgb]{ .651,  .651,  .651}\textbf{4.4772E+0} & 4.9874E+0 & 1.0352E+1 \\
\cline{2-6}\cline{8-12}          & PCC   & \cellcolor[rgb]{ .651,  .651,  .651}\textbf{9.8560E-1} & 9.7416E-1 & 9.8328E-1 & 2.8465E-1 &       & PCC   & 9.1451E-1 & \cellcolor[rgb]{ .651,  .651,  .651}\textbf{9.1843E-1} & 9.1196E-1 & 4.0436E-1 \\
\cline{2-6}\cline{8-12}          & SRCC  & 8.3050E-1 & 8.1905E-1 & \cellcolor[rgb]{ .651,  .651,  .651}\textbf{9.1239E-1} & 6.9996E-1 &       & SRCC  & 8.9866E-1 & 8.7992E-1 & \cellcolor[rgb]{ .651,  .651,  .651}\textbf{9.8220E-1} & 8.4578E-1 \\
    \hline
    \multirow{1}[6]{*}{F10} & RMSE  & \cellcolor[rgb]{ .651,  .651,  .651}\textbf{5.8167E+1} & 7.1668E+1 & 7.0629E+1 & 2.6264E+2 & \multirow{1}[6]{*}{F20} & RMSE  & 4.8658E-2 & 5.8129E-2 & \cellcolor[rgb]{ .651,  .651,  .651}\textbf{4.4185E-2} & 5.2539E-2 \\
\cline{2-6}\cline{8-12}          & PCC   & \cellcolor[rgb]{ .651,  .651,  .651}\textbf{9.8591E-1} & 9.7842E-1 & 9.8005E-1 & 6.9381E-1 &       & PCC   & 9.8499E-1 & 9.7863E-1 & \cellcolor[rgb]{ .651,  .651,  .651}\textbf{9.8761E-1} & 9.8251E-1 \\
\cline{2-6}\cline{8-12}          & SRCC  & 9.6279E-1 & 9.5453E-1 & \cellcolor[rgb]{ .651,  .651,  .651}\textbf{9.6651E-1} & 9.0516E-1 &       & SRCC  & 9.8416E-1 & 9.7760E-1 & \cellcolor[rgb]{ .651,  .651,  .651}\textbf{9.8710E-1} & 9.8196E-1 \\
    \hline
    \end{tabular}
}
\label{tab:metric_2D}
\end{table*}

\begin{table*}[htbp]
    \centering
    \caption{Comparisons of RMSE, PCC and SRCC obtained by four regression algorithms on benchmark problems $(d=10)$}
	\resizebox{\columnwidth}{!}{    
    \begin{tabular}{c|c|c|c|c|c||c|c|c|c|c|c}
    \hline
    \textbf{Problem} & \textbf{Metric} & \textbf{GP} & \textbf{RBFN} & \textbf{RF} & \textbf{SVR} & \textbf{Problem} & \textbf{Metric} & \textbf{GP} & \textbf{RBFN} & \textbf{RF} & \textbf{SVR} \\
    \hline
    \multirow{1}[6]{*}{F1} & RMSE  & \cellcolor[rgb]{ .651,  .651,  .651}\textbf{1.7299E+0} & 1.7636E+0 & 1.9689E+0 & 1.0508E+1 & \multirow{1}[6]{*}{F11} & RMSE  & \cellcolor[rgb]{ .651,  .651,  .651}\textbf{7.5782E+1} & 1.2276E+2 & 8.4930E+1 & 5.8035E+2 \\
\cline{2-6}\cline{8-12}          & PCC   & \cellcolor[rgb]{ .651,  .651,  .651}\textbf{9.9861E-1} & 9.9856E-1 & 9.9828E-1 & 9.5223E-1 &       & PCC   & \cellcolor[rgb]{ .651,  .651,  .651}\textbf{9.9431E-1} & 9.8604E-1 & 9.9342E-1 & 6.5587E-1 \\
\cline{2-6}\cline{8-12}          & SRCC  & 9.8381E-1 & 9.9032E-1 & \cellcolor[rgb]{ .651,  .651,  .651}\textbf{9.9443E-1} & 9.2279E-1 &       & SRCC  & 9.9059E-1 & 9.8960E-1 & \cellcolor[rgb]{ .651,  .651,  .651}\textbf{9.9059E-1} & 7.9249E-1 \\
    \hline
    \multirow{1}[6]{*}{F2} & RMSE  & \cellcolor[rgb]{ .651,  .651,  .651}\textbf{1.0182E+3} & 1.2786E+3 & 1.1599E+3 & 8.9599E+3 & \multirow{1}[6]{*}{F12} & RMSE  & \cellcolor[rgb]{ .651,  .651,  .651}\textbf{8.5225E+7} & 1.4132E+8 & 1.5177E+8 & 1.2503E+9 \\
\cline{2-6}\cline{8-12}          & PCC   & \cellcolor[rgb]{ .651,  .651,  .651}\textbf{9.9822E-1} & 9.9729E-1 & 9.9770E-1 & 8.5938E-1 &       & PCC   & \cellcolor[rgb]{ .651,  .651,  .651}\textbf{9.9739E-1} & 9.9281E-1 & 9.9270E-1 & 3.0347E-1 \\
\cline{2-6}\cline{8-12}          & SRCC  & 9.8394E-1 & 9.8794E-1 & \cellcolor[rgb]{ .651,  .651,  .651}\textbf{9.9256E-1} & 5.9707E-1 &       & SRCC  & 9.7218E-1 & 9.6638E-1 & \cellcolor[rgb]{ .651,  .651,  .651}\textbf{9.8772E-1} & 4.9073E-1 \\
    \hline
    \multirow{1}[6]{*}{F3} & RMSE  & 8.4958E+3 & 1.4600E+4 & \cellcolor[rgb]{ .651,  .651,  .651}\textbf{8.2164E+3} & 3.9288E+4 & \multirow{1}[6]{*}{F13} & RMSE  & \cellcolor[rgb]{ .651,  .651,  .651}\textbf{2.6751E+1} & 7.0974E+1 & 3.8205E+1 & 4.6031E+2 \\
\cline{2-6}\cline{8-12}          & PCC   & 9.9385E-1 & 9.8265E-1 & \cellcolor[rgb]{ .651,  .651,  .651}\textbf{9.9408E-1} & 8.6001E-1 &       & PCC   & \cellcolor[rgb]{ .651,  .651,  .651}\textbf{9.9854E-1} & 9.9019E-1 & 9.9725E-1 & 8.1323E-1 \\
\cline{2-6}\cline{8-12}          & SRCC  & 9.6171E-1 & 9.6198E-1 & \cellcolor[rgb]{ .651,  .651,  .651}\textbf{9.9279E-1} & 5.3054E-1 &       & SRCC  & 7.5720E-1 & 6.5339E-1 & \cellcolor[rgb]{ .651,  .651,  .651}\textbf{9.0342E-1} & 5.4879E-1 \\
    \hline
    \multirow{1}[6]{*}{F4} 		 & RMSE  & \cellcolor[rgb]{ .651,  .651,  .651}\textbf{6.0935E-1} & 6.3263E-1 & 8.1344E-1 & 2.2203E+0 & \multirow{1}[6]{*}{F14} & RMSE  & 3.4421E-2 & 6.1959E-2 & \cellcolor[rgb]{ .651,  .651,  .651}\textbf{2.8171E-2} & 6.4088E-2 \\
\cline{2-6}\cline{8-12}          & PCC   & \cellcolor[rgb]{ .651,  .651,  .651}\textbf{9.9604E-1} & 9.9575E-1 & 9.9319E-1 & 9.4638E-1 &       & PCC   & 9.3622E-1 & 8.1933E-1 & \cellcolor[rgb]{ .651,  .651,  .651}\textbf{9.5841E-1} & 7.7445E-1 \\
\cline{2-6}\cline{8-12}          & SRCC  & \cellcolor[rgb]{ .651,  .651,  .651}\textbf{9.9633E-1} & 9.9446E-1 & 9.9475E-1 & 9.5691E-1 &       & SRCC  & 9.4234E-1 & 8.1654E-1 & \cellcolor[rgb]{ .651,  .651,  .651}\textbf{9.6084E-1} & 8.0800E-1 \\
    \hline
    \multirow{1}[6]{*}{F5}       & RMSE  & 4.6539E-2 & 4.8853E-2 & \cellcolor[rgb]{ .651,  .651,  .651}\textbf{4.5680E-2} & 1.1754E-1 & \multirow{1}[6]{*}{F15} & RMSE  & \cellcolor[rgb]{ .651,  .651,  .651}\textbf{3.1974E+0} & 3.7984E+0 & 3.4427E+0 & 8.1132E+0 \\
\cline{2-6}\cline{8-12}          & PCC   & 9.7564E-1 & 9.7311E-1 & \cellcolor[rgb]{ .651,  .651,  .651}\textbf{9.7708E-1} & 8.3876E-1 &       & PCC   & \cellcolor[rgb]{ .651,  .651,  .651}\textbf{9.9203E-1} & 9.8875E-1 & 9.9118E-1 & 9.5124E-1 \\
\cline{2-6}\cline{8-12}          & SRCC  & 9.5180E-1 & 9.5233E-1 & \cellcolor[rgb]{ .651,  .651,  .651}\textbf{9.5597E-1} & 8.3469E-1 &       & SRCC  & 9.9141E-1 & 9.9001E-1 & \cellcolor[rgb]{ .651,  .651,  .651}\textbf{9.9205E-1} & 9.6200E-1 \\
    \hline
    \multirow{1}[6]{*}{F6}       & RMSE  & 4.0360E+0 & 5.4285E+0 & \cellcolor[rgb]{ .651,  .651,  .651}\textbf{3.8827E+0} & 1.4874E+1 & \multirow{1}[6]{*}{F16} & RMSE  & 4.4141E+0 & 4.7725E+0 & \cellcolor[rgb]{ .651,  .651,  .651}\textbf{4.2158E+0} & 1.2342E+1 \\
\cline{2-6}\cline{8-12}          & PCC   & 9.9643E-1 & 9.9355E-1 & \cellcolor[rgb]{ .651,  .651,  .651}\textbf{9.9676E-1} & 9.5154E-1 &       & PCC   & 9.9089E-1 & 9.8940E-1 & \cellcolor[rgb]{ .651,  .651,  .651}\textbf{9.9210E-1} & 9.3015E-1 \\
\cline{2-6}\cline{8-12}          & SRCC  & \cellcolor[rgb]{ .651,  .651,  .651}\textbf{9.9510E-1} & 9.9377E-1 & 9.9415E-1 & 9.1485E-1 &       & SRCC  & 9.9171E-1 & 9.9066E-1 & \cellcolor[rgb]{ .651,  .651,  .651}\textbf{9.9244E-1} & 9.3564E-1 \\
    \hline
    \multirow{1}[6]{*}{F7} & RMSE  & \cellcolor[rgb]{ .651,  .651,  .651}\textbf{3.1640E+2} & 3.2116E+2 & 4.0646E+2 & 3.9881E+3 & \multirow{1}[6]{*}{F17} & RMSE  & 3.6772E-1 & 3.9961E-1 & \cellcolor[rgb]{ .651,  .651,  .651}\textbf{3.1696E-1} & 4.5470E-1 \\
\cline{2-6}\cline{8-12}          & PCC   & \cellcolor[rgb]{ .651,  .651,  .651}\textbf{9.9671E-1} & 9.9663E-1 & 9.9537E-1 & 3.3030E-1 &       & PCC   & 9.8841E-1 & 9.8631E-1 & \cellcolor[rgb]{ .651,  .651,  .651}\textbf{9.9147E-1} & 9.8230E-1 \\
\cline{2-6}\cline{8-12}          & SRCC  & 9.8179E-1 & 9.8652E-1 & \cellcolor[rgb]{ .651,  .651,  .651}\textbf{9.9245E-1} & 4.6616E-1 &       & SRCC  & \cellcolor[rgb]{ .651,  .651,  .651}\textbf{9.8741E-1} & 9.8447E-1 & 9.8739E-1 & 9.8099E-1 \\
    \hline
    \multirow{1}[6]{*}{F8} & RMSE  & \cellcolor[rgb]{ .651,  .651,  .651}\textbf{4.2136E+2} & 4.9296E+2 & 5.0804E+2 & 4.5107E+3 & \multirow{1}[6]{*}{F18} & RMSE  & \cellcolor[rgb]{ .651,  .651,  .651}\textbf{3.1235E+3} & 3.6255E+3 & 4.0628E+3 & 3.3285E+4 \\
\cline{2-6}\cline{8-12}          & PCC   & \cellcolor[rgb]{ .651,  .651,  .651}\textbf{9.9603E-1} & 9.9456E-1 & 9.9450E-1 & 4.6619E-1 &       & PCC   & \cellcolor[rgb]{ .651,  .651,  .651}\textbf{9.9533E-1} & 9.9380E-1 & 9.9284E-1 & 4.0551E-1 \\
\cline{2-6}\cline{8-12}          & SRCC  & 9.9165E-1 & 9.9146E-1 & \cellcolor[rgb]{ .651,  .651,  .651}\textbf{9.9396E-1} & 5.0729E-1 &       & SRCC  & \cellcolor[rgb]{ .651,  .651,  .651}\textbf{9.9208E-1} & 9.9109E-1 & 9.9188E-1 & 5.7869E-1 \\
    \hline
    \multirow{1}[6]{*}{F9} & RMSE  & \cellcolor[rgb]{ .651,  .651,  .651}\textbf{3.1078E+6} & 3.9717E+6 & 4.8207E+6 & 3.5654E+7 & \multirow{1}[6]{*}{F19} & RMSE  & \cellcolor[rgb]{ .651,  .651,  .651}\textbf{3.3832E+0} & 4.1114E+0 & 4.2199E+0 & 1.0164E+1 \\
\cline{2-6}\cline{8-12}          & PCC   & \cellcolor[rgb]{ .651,  .651,  .651}\textbf{9.9601E-1} & 9.9335E-1 & 9.9111E-1 & 3.6766E-1 &       & PCC   & \cellcolor[rgb]{ .651,  .651,  .651}\textbf{9.7775E-1} & 9.6694E-1 & 9.6695E-1 & 7.9914E-1 \\
\cline{2-6}\cline{8-12}          & SRCC  & 9.8857E-1 & 9.8852E-1 & \cellcolor[rgb]{ .651,  .651,  .651}\textbf{9.9138E-1} & 5.1935E-1 &       & SRCC  & 9.7974E-1 & 9.7945E-1 & \cellcolor[rgb]{ .651,  .651,  .651}\textbf{9.9234E-1} & 9.4797E-1 \\
    \hline
    \multirow{1}[6]{*}{F10} & RMSE  & \cellcolor[rgb]{ .651,  .651,  .651}\textbf{4.7094E+2} & 6.1865E+2 & 5.5723E+2 & 4.6398E+3 & \multirow{1}[6]{*}{F20} & RMSE  & 6.3115E-2 & 7.6011E-2 & \cellcolor[rgb]{ .651,  .651,  .651}\textbf{4.9801E-2} & 1.1880E-1 \\
\cline{2-6}\cline{8-12}          & PCC   & \cellcolor[rgb]{ .651,  .651,  .651}\textbf{9.9545E-1} & 9.9220E-1 & 9.9390E-1 & 5.1505E-1 &       & PCC   & 9.8540E-1 & 9.7913E-1 & \cellcolor[rgb]{ .651,  .651,  .651}\textbf{9.9116E-1} & 9.4940E-1 \\
\cline{2-6}\cline{8-12}          & SRCC  & 9.9190E-1 & 9.9131E-1 & \cellcolor[rgb]{ .651,  .651,  .651}\textbf{9.9376E-1} & 5.3709E-1 &       & SRCC  & 9.7823E-1 & 9.7534E-1 & \cellcolor[rgb]{ .651,  .651,  .651}\textbf{9.8558E-1} & 9.8114E-1 \\
    \hline
    \end{tabular}
}
\label{tab:metric_10D}
\end{table*}

\begin{table*}[htbp]
    \centering
    \caption{Comparisons of RMSE, PCC and SRCC obtained by four regression algorithms on benchmark problems $(d=30)$}
    \resizebox{\columnwidth}{!}{
    \begin{tabular}{c|c|c|c|c|c||c|c|c|c|c|c}
    \hline
    \textbf{Problem} & \textbf{Metric} & \textbf{GP} & \textbf{RBFN} & \textbf{RF} & \textbf{SVR} & \textbf{Problem} & \textbf{Metric} & \textbf{GP} & \textbf{RBFN} & \textbf{RF} & \textbf{SVR} \\
    \hline
    \multirow{1}[6]{*}{F1} & RMSE  & \cellcolor[rgb]{ .651,  .651,  .651}\textbf{3.1664E+0} & 4.3754E+0 & 5.9464E+0 & 2.6140E+1 & \multirow{1}[6]{*}{F11} & RMSE  & 6.1026E+2 & \cellcolor[rgb]{ .651,  .651,  .651}\textbf{5.7617E+2} & 7.6408E+2 & 7.1468E+3 \\
\cline{2-6}\cline{8-12}          & PCC   & \cellcolor[rgb]{ .651,  .651,  .651}\textbf{9.9970E-1} & 9.9942E-1 & 9.9894E-1 & 9.8077E-1 &       & PCC   & 9.9656E-1 & \cellcolor[rgb]{ .651,  .651,  .651}\textbf{9.9696E-1} & 9.9484E-1 & 4.9608E-1 \\
\cline{2-6}\cline{8-12}          & SRCC  & \cellcolor[rgb]{ .651,  .651,  .651}\textbf{9.9955E-1} & 9.9933E-1 & 9.9832E-1 & 9.6490E-1 &       & SRCC  & 9.9691E-1 & \cellcolor[rgb]{ .651,  .651,  .651}\textbf{9.9696E-1} & 9.9399E-1 & 6.4670E-1 \\
    \hline
    \multirow{1}[6]{*}{F2} & RMSE  & \cellcolor[rgb]{ .651,  .651,  .651}\textbf{7.3927E+3} & 8.1327E+3 & 1.1700E+4 & 2.9682E+5 & \multirow{1}[6]{*}{F12} & RMSE  & \cellcolor[rgb]{ .651,  .651,  .651}\textbf{8.5205E+8} & 1.0921E+9 & 1.4697E+9 & 1.3268E+10 \\
\cline{2-6}\cline{8-12}          & PCC   & \cellcolor[rgb]{ .651,  .651,  .651}\textbf{9.9967E-1} & 9.9960E-1 & 9.9918E-1 & 7.8920E-1 &       & PCC   & \cellcolor[rgb]{ .651,  .651,  .651}\textbf{9.9766E-1} & 9.9618E-1 & 9.9369E-1 & 1.4329E-1 \\
\cline{2-6}\cline{8-12}          & SRCC  & \cellcolor[rgb]{ .651,  .651,  .651}\textbf{9.9949E-1} & 9.9937E-1 & 9.9847E-1 & 8.1683E-1 &       & SRCC  & 9.8387E-1 & 9.8806E-1 & \cellcolor[rgb]{ .651,  .651,  .651}\textbf{9.9281E-1} & 1.7774E-1 \\
    \hline
    \multirow{1}[6]{*}{F3} & RMSE  & \cellcolor[rgb]{ .651,  .651,  .651}\textbf{4.5734E+4} & 6.8867E+4 & 5.7877E+4 & 1.1203E+6 & \multirow{1}[6]{*}{F13} & RMSE  & 1.6142E+2 & 2.6088E+2 & \cellcolor[rgb]{ .651,  .651,  .651}\textbf{1.4616E+2} & 1.9077E+3 \\
\cline{2-6}\cline{8-12}          & PCC   & \cellcolor[rgb]{ .651,  .651,  .651}\textbf{9.9909E-1} & 9.9792E-1 & 9.9855E-1 & 7.5709E-1 &       & PCC   & 9.9608E-1 & 9.8951E-1 & \cellcolor[rgb]{ .651,  .651,  .651}\textbf{9.9675E-1} & 7.3189E-1 \\
\cline{2-6}\cline{8-12}          & SRCC  & \cellcolor[rgb]{ .651,  .651,  .651}\textbf{9.9906E-1} & 9.9806E-1 & 9.9804E-1 & 8.4369E-1 &       & SRCC  & 8.5153E-1 & 7.8392E-1 & \cellcolor[rgb]{ .651,  .651,  .651}\textbf{9.4461E-1} & 6.0091E-1 \\
    \hline
    \multirow{1}[6]{*}{F4} & RMSE  & \cellcolor[rgb]{ .651,  .651,  .651}\textbf{1.4897E-1} & 3.7718E-1 & 3.7555E-1 & 1.3550E+0 & \multirow{1}[6]{*}{F14} & RMSE  & 2.5218E-2 & 8.6535E-2 & \cellcolor[rgb]{ .651,  .651,  .651}\textbf{2.1899E-2} & 7.9183E-2 \\
\cline{2-6}\cline{8-12}          & PCC   & \cellcolor[rgb]{ .651,  .651,  .651}\textbf{9.9874E-1} & 9.9183E-1 & 9.9204E-1 & 8.9427E-1 &       & PCC   & 9.7414E-1 & 7.8203E-1 & \cellcolor[rgb]{ .651,  .651,  .651}\textbf{9.8112E-1} & 7.1034E-1 \\
\cline{2-6}\cline{8-12}          & SRCC  & \cellcolor[rgb]{ .651,  .651,  .651}\textbf{9.9888E-1} & 9.7018E-1 & 9.9810E-1 & 9.3530E-1 &       & SRCC  & 9.5246E-1 & 7.4453E-1 & \cellcolor[rgb]{ .651,  .651,  .651}\textbf{9.6295E-1} & 4.7142E-1 \\
    \hline
    \multirow{1}[6]{*}{F5} & RMSE  & \cellcolor[rgb]{ .651,  .651,  .651}\textbf{5.8351E-2} & 1.0707E-1 & 8.1513E-2 & 3.1414E-1 & \multirow{1}[6]{*}{F15} & RMSE  & \cellcolor[rgb]{ .651,  .651,  .651}\textbf{1.1107E+1} & 1.7293E+1 & 1.3455E+1 & 7.1742E+1 \\
\cline{2-6}\cline{8-12}          & PCC   & \cellcolor[rgb]{ .651,  .651,  .651}\textbf{9.9892E-1} & 9.9641E-1 & 9.9793E-1 & 9.6869E-1 &       & PCC   & \cellcolor[rgb]{ .651,  .651,  .651}\textbf{9.9482E-1} & 9.8756E-1 & 9.9271E-1 & 7.6305E-1 \\
\cline{2-6}\cline{8-12}          & SRCC  & 9.9664E-1 & 9.9348E-1 & \cellcolor[rgb]{ .651,  .651,  .651}\textbf{9.9705E-1} & 9.7934E-1 &       & SRCC  & \cellcolor[rgb]{ .651,  .651,  .651}\textbf{9.9433E-1} & 9.9250E-1 & 9.9274E-1 & 7.6027E-1 \\
    \hline
    \multirow{1}[6]{*}{F6} & RMSE  & \cellcolor[rgb]{ .651,  .651,  .651}\textbf{6.4422E+0} & 1.2316E+1 & 8.1210E+0 & 4.0464E+1 & \multirow{1}[6]{*}{F16} & RMSE  & \cellcolor[rgb]{ .651,  .651,  .651}\textbf{1.4891E+1} & 1.9211E+1 & 1.9092E+1 & 1.1678E+2 \\
\cline{2-6}\cline{8-12}          & PCC   & \cellcolor[rgb]{ .651,  .651,  .651}\textbf{9.9896E-1} & 9.9624E-1 & 9.9837E-1 & 9.6152E-1 &       & PCC   & \cellcolor[rgb]{ .651,  .651,  .651}\textbf{9.9520E-1} & 9.9201E-1 & 9.9250E-1 & 6.5834E-1 \\
\cline{2-6}\cline{8-12}          & SRCC  & \cellcolor[rgb]{ .651,  .651,  .651}\textbf{9.9883E-1} & 9.9741E-1 & 9.9851E-1 & 9.6091E-1 &       & SRCC  & \cellcolor[rgb]{ .651,  .651,  .651}\textbf{9.9588E-1} & 9.9333E-1 & 9.9332E-1 & 6.9728E-1 \\
    \hline
    \multirow{1}[6]{*}{F7} & RMSE  & \cellcolor[rgb]{ .651,  .651,  .651}\textbf{1.8187E+3} & 2.5706E+3 & 2.1993E+3 & 2.2890E+4 & \multirow{1}[6]{*}{F17} & RMSE  & \cellcolor[rgb]{ .651,  .651,  .651}\textbf{1.2600E+0} & 2.8985E+0 & 1.5206E+0 & 4.4456E+0 \\
\cline{2-6}\cline{8-12}          & PCC   & \cellcolor[rgb]{ .651,  .651,  .651}\textbf{9.9671E-1} & 9.9367E-1 & 9.9540E-1 & 4.4227E-1 &       & PCC   & \cellcolor[rgb]{ .651,  .651,  .651}\textbf{9.8696E-1} & 9.3534E-1 & 9.8134E-1 & 8.2768E-1 \\
\cline{2-6}\cline{8-12}          & SRCC  & 9.9188E-1 & 9.9305E-1 & \cellcolor[rgb]{ .651,  .651,  .651}\textbf{9.9386E-1} & 5.9837E-1 &       & SRCC  & \cellcolor[rgb]{ .651,  .651,  .651}\textbf{9.7828E-1} & 9.6772E-1 & 9.7314E-1 & 8.5502E-1 \\
    \hline
    \multirow{1}[6]{*}{F8} & RMSE  & 2.4025E+3 & \cellcolor[rgb]{ .651,  .651,  .651}\textbf{2.3311E+3} & 3.1974E+3 & 2.6548E+4 & \multirow{1}[6]{*}{F18} & RMSE  & 4.8301E+4 & 5.8401E+4 & \cellcolor[rgb]{ .651,  .651,  .651}\textbf{4.6411E+4} & 5.1075E+5 \\
\cline{2-6}\cline{8-12}          & PCC   & 9.9597E-1 & \cellcolor[rgb]{ .651,  .651,  .651}\textbf{9.9620E-1} & 9.9333E-1 & 5.8200E-1 &       & PCC   & 9.9530E-1 & 9.9312E-1 & \cellcolor[rgb]{ .651,  .651,  .651}\textbf{9.9585E-1} & 4.6182E-1 \\
\cline{2-6}\cline{8-12}          & SRCC  & 9.9596E-1 & \cellcolor[rgb]{ .651,  .651,  .651}\textbf{9.9608E-1} & 9.9406E-1 & 6.2086E-1 &       & SRCC  & 9.9473E-1 & \cellcolor[rgb]{ .651,  .651,  .651}\textbf{9.9510E-1} & 9.9470E-1 & 6.4254E-1 \\
    \hline
    \multirow{1}[6]{*}{F9} & RMSE  & \cellcolor[rgb]{ .651,  .651,  .651}\textbf{1.9825E+7} & 2.6982E+7 & 3.8388E+7 & 3.1390E+8 & \multirow{1}[6]{*}{F19} & RMSE  & \cellcolor[rgb]{ .651,  .651,  .651}\textbf{7.5569E+1} & 9.6097E+1 & 7.5740E+1 & 7.0118E+2 \\
\cline{2-6}\cline{8-12}          & PCC   & \cellcolor[rgb]{ .651,  .651,  .651}\textbf{9.9790E-1} & 9.9611E-1 & 9.9251E-1 & 1.8441E-1 &       & PCC   & 9.9469E-1 & 9.9170E-1 & \cellcolor[rgb]{ .651,  .651,  .651}\textbf{9.9478E-1} & 6.1791E-1 \\
\cline{2-6}\cline{8-12}          & SRCC  & \cellcolor[rgb]{ .651,  .651,  .651}\textbf{9.9316E-1} & 9.9320E-1 & 9.9271E-1 & 3.9600E-1 &       & SRCC  & 9.7199E-1 & \cellcolor[rgb]{ .651,  .651,  .651}\textbf{9.8092E-1} & 9.9193E-1 & 8.1939E-1 \\
    \hline
    \multirow{1}[6]{*}{F10} & RMSE  & \cellcolor[rgb]{ .651,  .651,  .651}\textbf{2.3746E+3} & 2.6324E+3 & 3.1720E+3 & 2.5763E+4 & \multirow{1}[6]{*}{F20} & RMSE  & \cellcolor[rgb]{ .651,  .651,  .651}\textbf{9.4363E-2} & 1.8844E-1 & 1.1060E-1 & 3.0212E-1 \\
\cline{2-6}\cline{8-12}          & PCC   & \cellcolor[rgb]{ .651,  .651,  .651}\textbf{9.9576E-1} & 9.9480E-1 & 9.9279E-1 & 5.8482E-1 &       & PCC   & \cellcolor[rgb]{ .651,  .651,  .651}\textbf{9.8325E-1} & 9.3592E-1 & 9.7715E-1 & 8.1226E-1 \\
\cline{2-6}\cline{8-12}          & SRCC  & \cellcolor[rgb]{ .651,  .651,  .651}\textbf{9.9487E-1} & 9.9380E-1 & 9.9214E-1 & 6.5200E-1 &       & SRCC  & 9.7602E-1 & 9.4643E-1 & \cellcolor[rgb]{ .651,  .651,  .651}\textbf{9.7782E-1} & 8.3298E-1 \\
    \hline
    \end{tabular}
}
\label{tab:metric_30D}
\end{table*}

To have a better understanding of the generalisation performance of different surrogate models (especially the relationship between the predicted performance and its ground truth given a particular parameter configuration), we calculate the Pearson correlation coefficient (PCC) of the results:
\begin{equation}
    PCC=\frac{\mathtt{cov}(X,Y)}{\sigma_X\sigma_Y}
\label{eq:pcc}
\end{equation}
where $X$ represents the set of observed approximation errors of all parameter configurations in the testing set while $Y$ is the set of approximation errors estimated by a surrogate model. $\mathtt{cov}(X,Y)$ is the covariance of $X$ and $Y$, $\sigma_X$ and $\sigma_Y$ are the standard deviations of $X$ and $Y$. In particular, a higher PCC indicates a better correlation between the predicted performance and the ground truth. 

\begin{figure}[htbp]
    \centering
    \includegraphics[width=\linewidth]{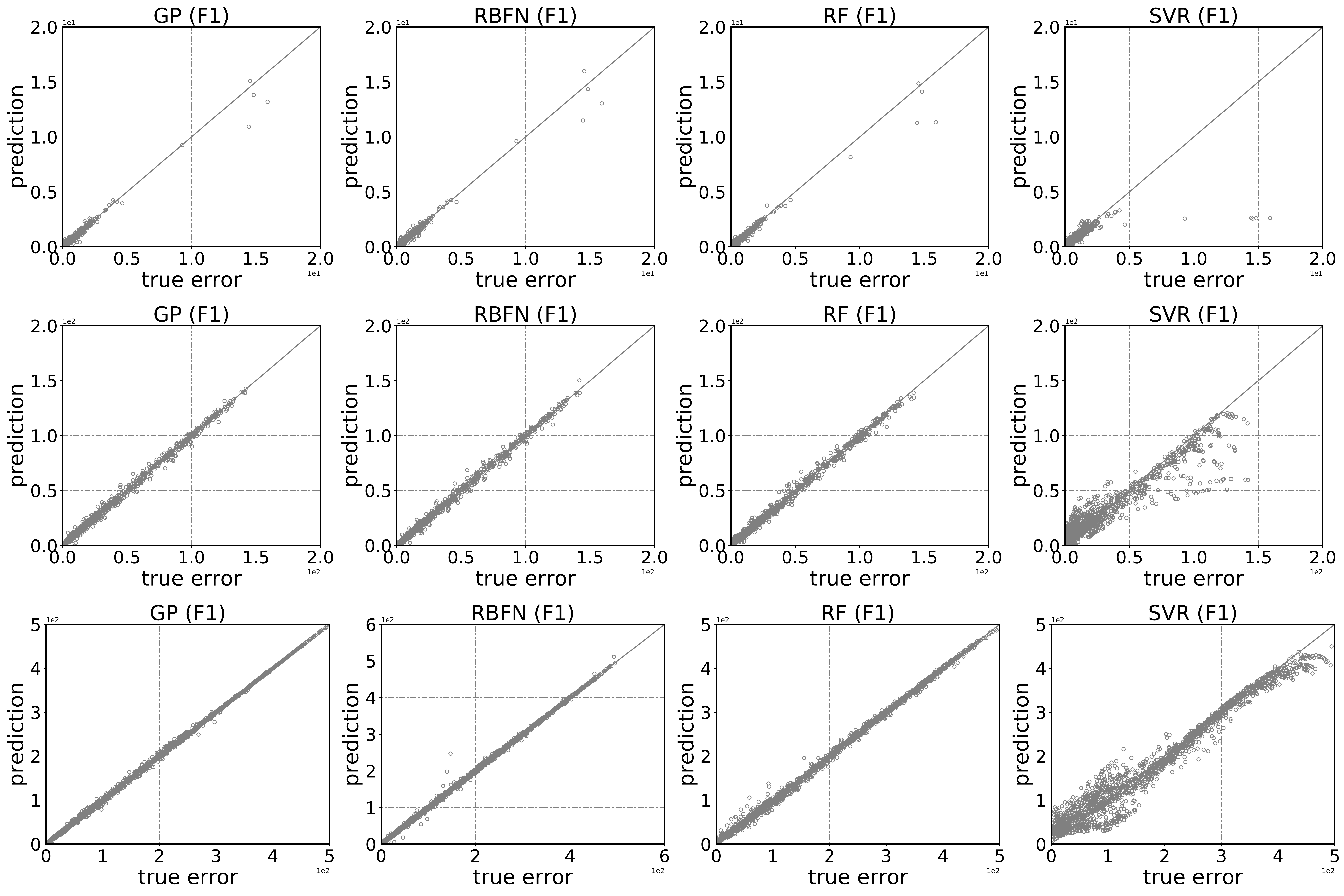}
    \caption{Scatter plots of the empirical performance predicted by a surrogate model \textit{vs} the observed empirical performance on the testing set (i.e. unseen parameter configurations). In particular, three rows respectively represent results on F1 where $d=2,10,30$.}
    \label{fig:F1}
\end{figure}

\begin{figure}[htbp]
    \centering
    \includegraphics[width=\linewidth]{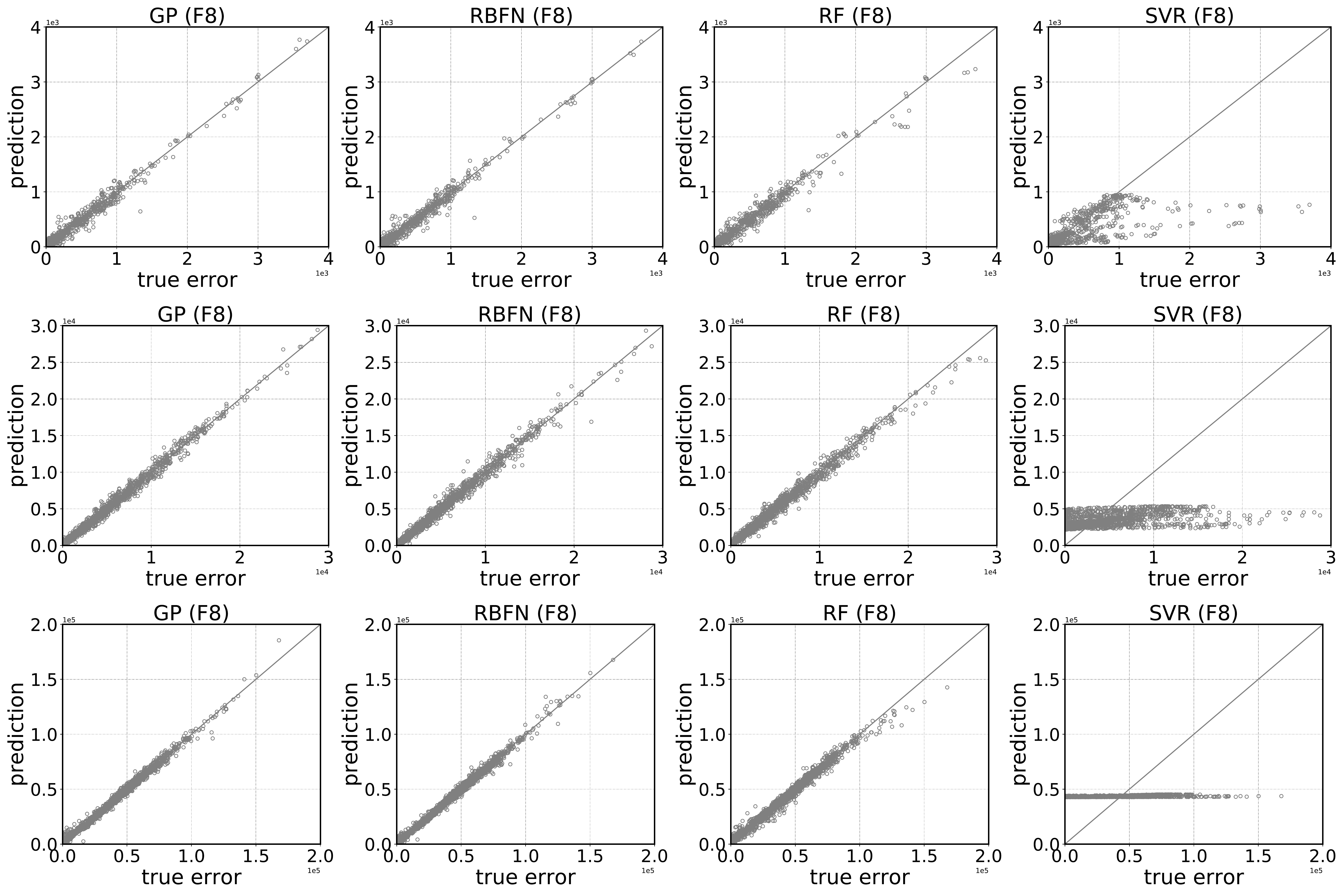}
    \caption{Scatter plots of the empirical performance predicted by a surrogate model \textit{vs} the observed empirical performance on the testing set (i.e. unseen parameter configurations). In particular, three rows respectively represent results on F8 where $d=2,10,30$.}
    \label{fig:F8}
\end{figure}

\begin{figure}[htbp]
    \centering
    \includegraphics[width=\linewidth]{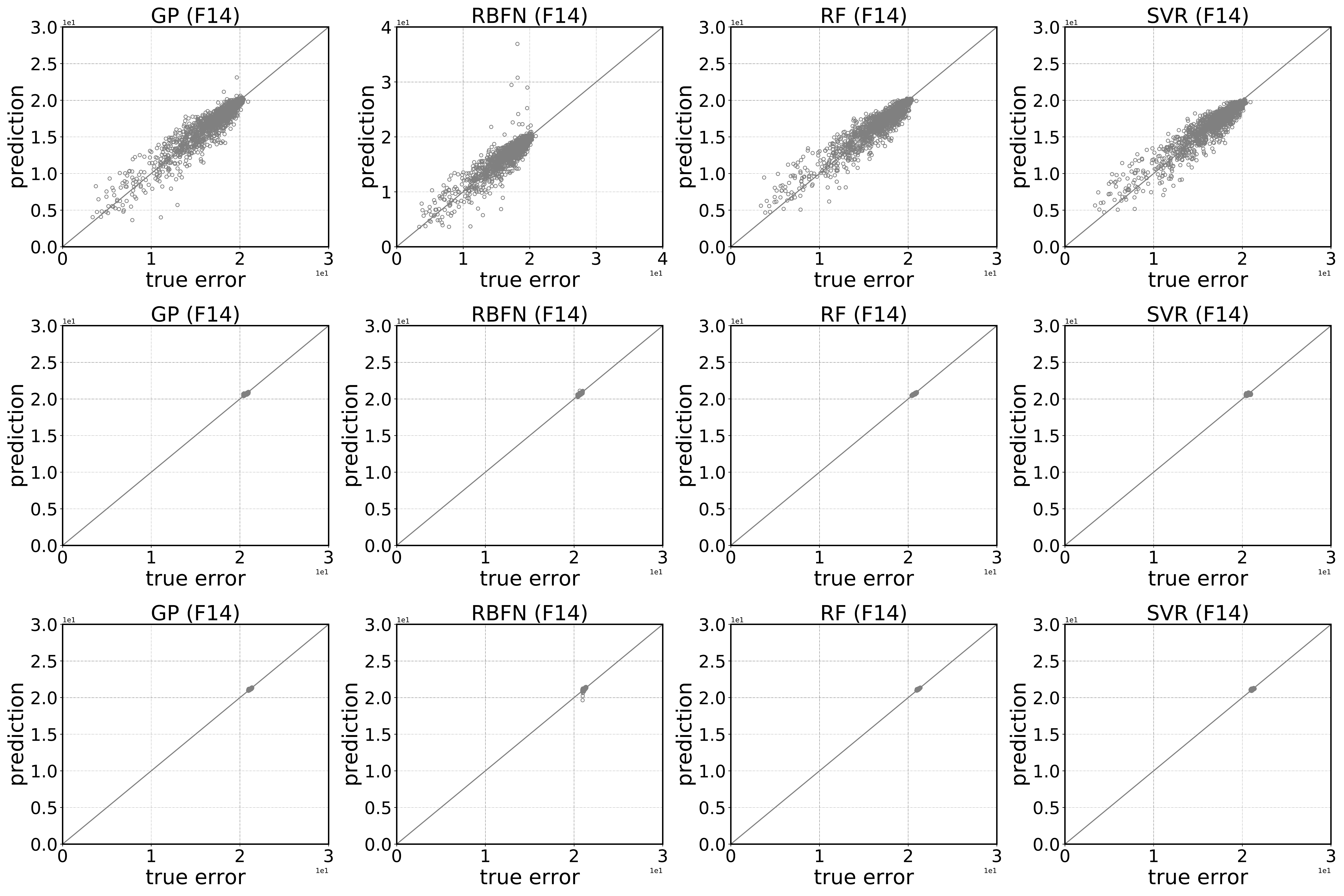}
    \caption{Scatter plots of the empirical performance predicted by a surrogate model \textit{vs} the observed empirical performance on the testing set (i.e. unseen parameter configurations). In particular, three rows respectively represent results on F14 where $d=2,10,30$.}
    \label{fig:F14}
\end{figure}

From the results shown in Tables~\ref{tab:metric_2D} to~\ref{tab:metric_30D}, we can see that the observations are in line with the RMSE. The performance of GP and RF are the most competitive regression algorithms in almost all cases, where the correlation between the predicted performance and its ground truth is relatively high. The performance of RBFN is very close to those of GP and RF, while the PCC obtained by SVR is the worst. To have a visual understanding of this point, we also provide the scatter plots of \textit{ground truth vs predicted performance} in Figures~\ref{fig:F1} to \ref{fig:F14}\footnote{More comprehensive figures are moved to the supplementary document, which can be downloaded from \href{https://coda-group.github.io/cec19-supp.pdf}{http://coda-group.github.io/cec19-supp.pdf}.}. According to the observations from these figures and Tables~\ref{tab:metric_2D} to~\ref{tab:metric_30D}, we summarise our findings as follows. 
\begin{itemize}
    \item As shown in Tables~\ref{tab:metric_2D} to~\ref{tab:metric_30D}, the RMSEs of all four regression algorithms are huge (over $10^7$) on F9 and F12. This is because the performance of DE are miserable on these two test problems with almost all sampled 5,940 parameter configurations. Accordingly, the deviations of the predicted empirical performance are in a relatively large scale. This also explains the increase of RMSEs with the problem dimensionality. However, according to PCCs, we find that the correlation between the predicted empirical performance and the ground truth of GP, RBFN and RF are acceptable.
    \item The RMSEs of the first six elementary test problems (i.e. F1 to F6), which are relatively simple, are better than those from CEC 2005 competition. Accordingly, the deviations between the predicted performance and the ground truth are small. This indicates that most parameter configurations are able to lead to an acceptable performance of DE. In other words, DE is not sensitive to its configurations on these problems.
    \item As shown in~\pref{fig:F8}, we find that SVR largely underestimates the approximation error on F8. Similar observations can be found on F7, F9, F10, F12 and F18 as shown in the supplementary document.
    \item As shown in~\pref{fig:F14}, we find that scatter plots are crowded in the middle region of the diagonal line. This implies that all parameter configurations fail to lead to a decent result. Similar observations can be found on F13 and F20 when the number of variables becomes large in the supplementary document.
\end{itemize}

Based on the above discussions, we come up with the following response to RQ1:

\vspace{0.5em}
\noindent
\framebox{\parbox{\dimexpr\linewidth-2\fboxsep-2\fboxrule}{
    \underline{\textbf{Response to RQ1}}: \textbf{\textit{GP and RF are the best regression algorithms for building the surrogate model of empirical performance. In addition, the quality of the surrogate model depend on the quality of the performance data.}}
}}

\subsection{Comparisons of Performance Ranks Obtained by Different Surrogate Models}
\label{sec:rank}

When using a surrogate in a sequential model-based PO, the prediction accuracy of this model is not utterly important. Instead, reliably differentiating the promising ones with respect to their unpromising counterparts can also provide useful information to guide the optimisation process. In other words, for a set of parameter configurations, we expect that the ranks (or the order) of the empirical performance predicted by a surrogate model can follow those of the ground truth. To this end, we consider using the Spearman's rank correlation coefficient (SRCC) to measure the statistical dependence between the ranks of the predicted performance and the ground truth. Note that the calculation of SRCC is almost the same as that of PCC, except that the raw data is replaced by the corresponding ranks.
\begin{equation}
SRCC=\frac{\mathtt{cov}(r_X,r_Y)}{\sigma_{r_X}\sigma_{r_Y}}
\label{eq:srcc}
\end{equation}
where $r_X$ indicates the ranks of the observed approximation errors of all parameters configurations in the testing set while $r_Y$ is the ranks of those estimated approximation errors. A higher SRCC indicates a better dependency between the predicted performance and the ground truth.

From the results shown in Tables~\ref{tab:metric_2D} to~\ref{tab:metric_30D}, we can still come up with the conclusion that GP and RF are the most reliable regression algorithms for building the surrogate model of the empirical performance. They almost dominate the top two positions in terms of SRCC. It is interesting to note that the SRCCs obtained by SVR are not as poor as its performance on RMSE and PCC. It is even comparable with GP and RF in some cases, e.g. on F20. This suggests that the prediction made by SVR has a decent chance to differentiate the order between two parameter configurations. In this case, SVR might be useful in a model-based PO process where it can be used as a comparison-based surrogate~\cite{LoshchilovSS10}. Furthermore, we also notice that RBFN does not show a good performance on SRCC. It is even sometimes worse than SVR. This indicates that although the prediction made by RBFN is numerically close to the ground truth, it may still mislead a model-based PO as it messes up the order of similar parameter configurations.

Based on the above discussion, we come up with the following response to RQ2:

\vspace{0.5em}
\noindent
\framebox{\parbox{\dimexpr\linewidth-2\fboxsep-2\fboxrule}{
    \underline{\textbf{Response to RQ2}}: \textbf{\textit{GP and RF are able to preserve the order of the empirical performance of different parameter configurations. In particular, SVR, which performs poorly on predicting the empirical performance, shows comparable performance for order preservation.}}
}}

\subsection{Comparisons of Landscape Approximation}
\label{sec:landscape}

\begin{figure}[htbp]
    \centering
    \includegraphics[width=\linewidth]{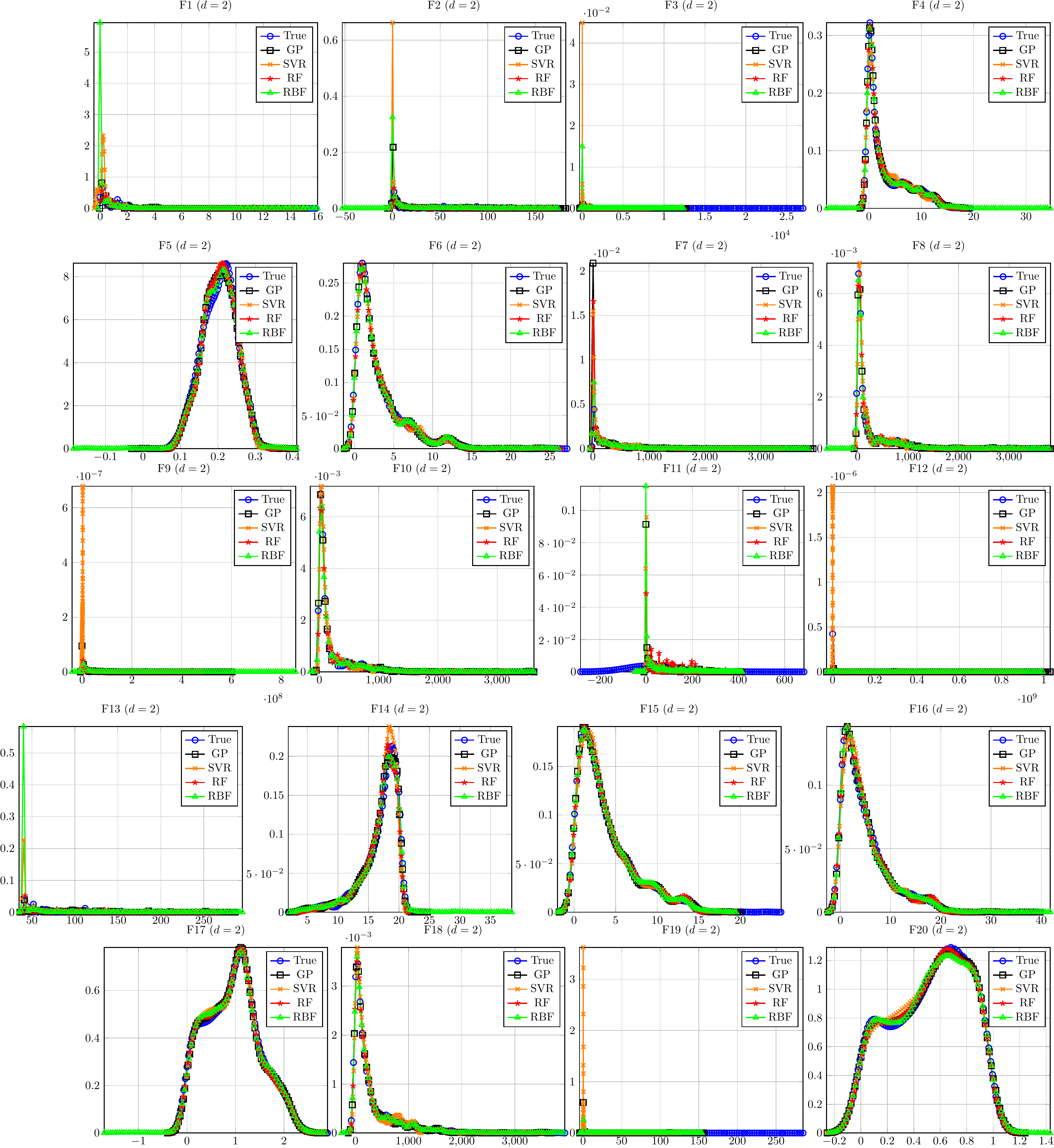}
    \caption{Estimated probability density distribution of the empirical performance predicted by four different regression algorithms and the ground truth ($d=2$).}
    \label{fig:KL_2D}
\end{figure}

\begin{figure}[htbp]
    \centering
    \includegraphics[width=\linewidth]{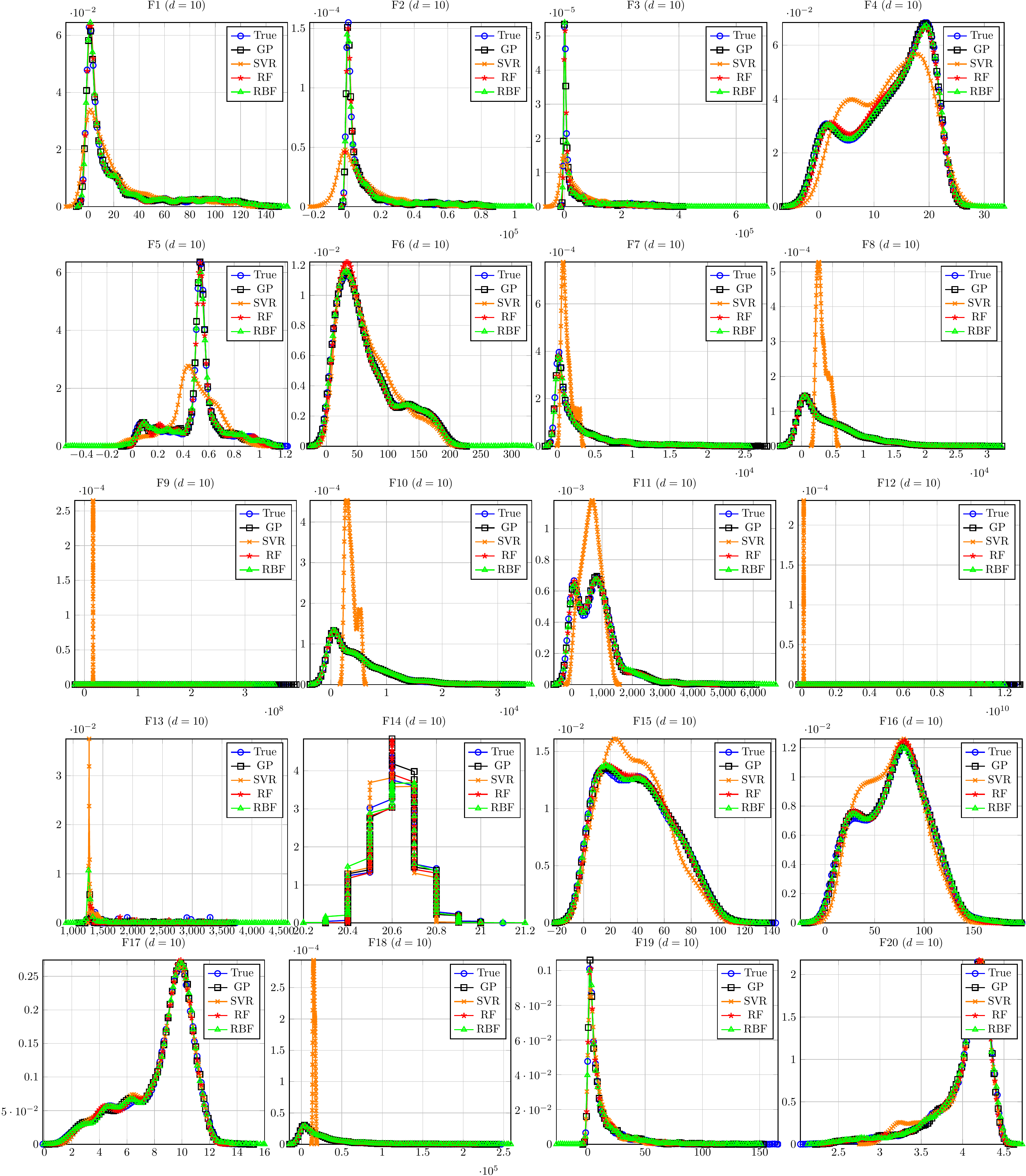}
    \caption{Estimated probability density distribution of the empirical performance predicted by four different regression algorithms and the ground truth ($d=10$).}
    \label{fig:KL_10D}
\end{figure}

\begin{figure}[htbp]
    \centering
    \includegraphics[width=\linewidth]{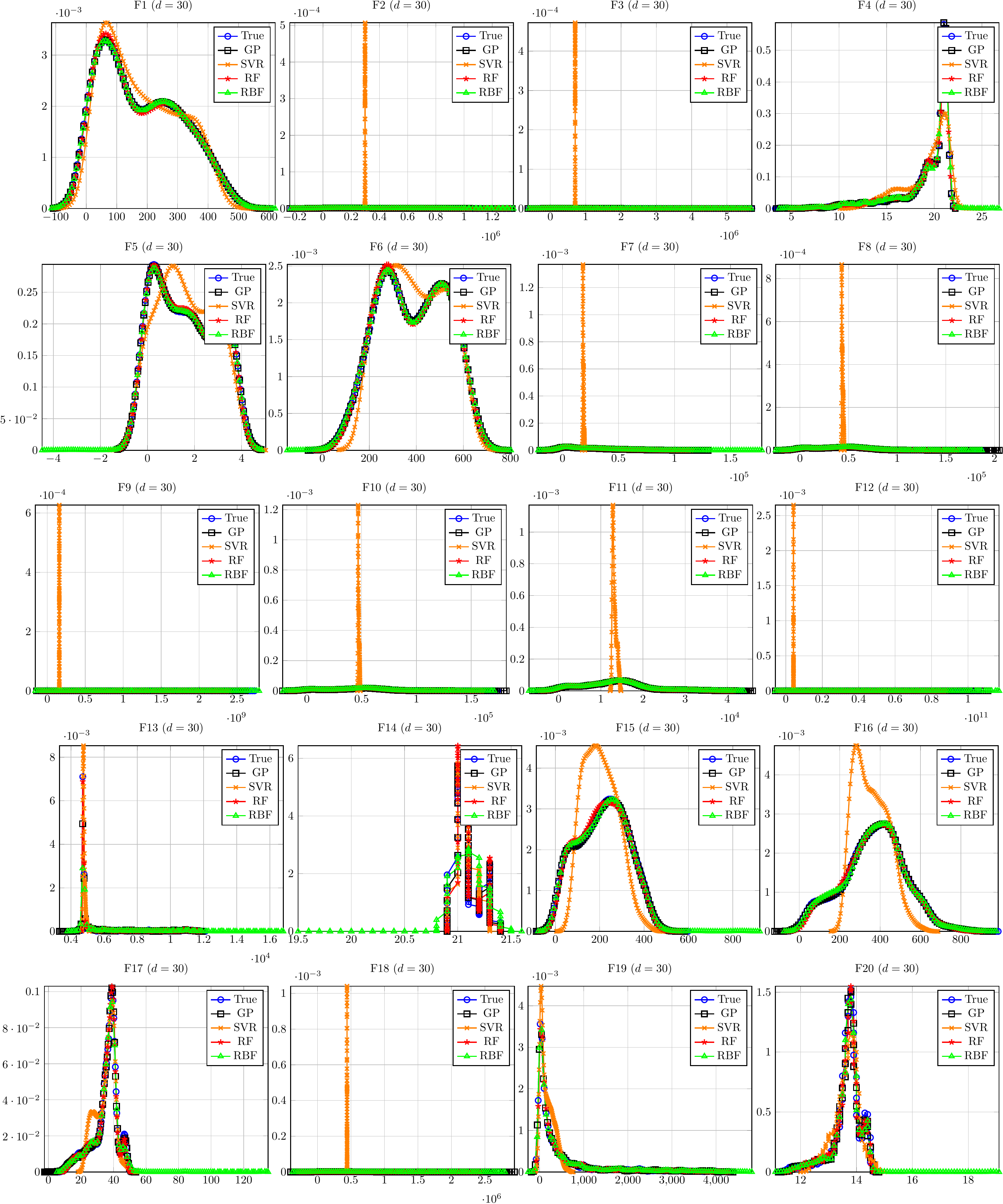}
    \caption{Estimated probability density distribution of the empirical performance predicted by four different regression algorithms and the ground truth ($d=30$).}
    \label{fig:KL_30D}
\end{figure}

In previous subsections, we mainly focus on investigating the quality of surrogate models from the approximation accuracy perspective. For the last RQ, we plan to study of the quality of surrogate models from a landscape analysis perspective. Considering the testing data set, we compare the landscapes of the empirical performance predicted by different regression algorithms to the landscape of the ground truth. To this end, we use the kernel density estimation (KDE) method\footnote{https://uk.mathworks.com/help/stats/ksdensity.html} to estimate a probability density function (PDF) of the empirical performance. To have a visual comparison, Figs~\ref{fig:KL_2D} to~\ref{fig:KL_30D} shows the plots of the estimated PDFs of four different regression algorithms and the ground truth. From these figures, we can see that the prediction made by GP, RF and RBFN almost fit the distribution of the ground truth. In contrast, the estimated PDF of SVR deviates from the ground truth in many cases. This becomes more evident when the dimensionality of the underlying problem becomes large.

Since the surrogate model considered in this paper is a mapping between a parameter configuration and its corresponding empirical performance, it is interesting to consider a more complex landscape that is a joint probability distribution of parameter configuration and empirical performance. As it is non-trivial to visualise a multi-dimensional distribution, we try to understand the proximity of the landscape approximated by the surrogate model and that of the ground truth from a statistical distance perspective. To this end, we apply the earth mover's distance (EMD)~\cite{RubnerTG00}, also known as Wasserstein metric, to evaluate the dissimilarity between two multi-dimensional distributions. Generally speaking, given two distributions, the EMD measures the minimum cost of turning one distribution into the other. In our context, similar landscapes are expected to have a relatively small EMD whereas large EMD values will imply that the landscapes are significantly different from each other. Due to the page limit, we do not intend to elaborate the calculation procedure of EMD, interested readers can refer to~\cite{RubnerTG00} for more details. From the comparison results of EMD values shown in~\pref{tab:emd}, we find that GP, RF and RBFN have the same level of approximation to the ground truth whereas the divergence values obtained by SVR are relatively large in almost all cases. All these observations are also in line with the RMSEs discussed in~\pref{sec:RMSE_comparison}.

Based on the above discussion, we come up with the following response to RQ3:

\vspace{0.5em}
\noindent
\framebox{\parbox{\dimexpr\linewidth-2\fboxsep-2\fboxrule}{
    \underline{\textbf{Response to RQ3}}: \textbf{\textit{The landscapes of the empirical performance predicted GP, RF and RBFN well approximate the ground truth; while the landscapes obtained by SVR deviate from the ground truth to a certain extent.}}
}}

\begin{table*}[htbp]
    \centering
    \caption{Comparisons of EMD between the surrogate model built by four regression algorithms and the ground truth}
    \resizebox{\columnwidth}{!}{    
    \begin{tabular}{c|c|c|c|c|c||c|c|c|c|c|c}
    \hline
    \textbf{Problem} & \textbf{$d$} & \textbf{GP} & \textbf{RBFN} & \textbf{RF} & \textbf{SVR} & \textbf{Problem} & \textbf{$d$} & \textbf{GP} & \textbf{RBFN} & \textbf{RF} & \textbf{SVR} \\
    \hline
    \multirow{1}[6]{*}{F1} & 2     & 3.9123E-2 & 4.1131E-2 & \cellcolor[rgb]{ .651,  .651,  .651}\textbf{3.8218E-2} & 2.1449E-1 & \multirow{1}[6]{*}{F11} & 2     & \cellcolor[rgb]{ .651,  .651,  .651}\textbf{1.6881E+0} & 2.0925E+0 & 2.1808E+0 & 1.0413E+1 \\
\cline{2-6}\cline{8-12}          & 10    & 7.4359E-1 & \cellcolor[rgb]{ .651,  .651,  .651}\textbf{7.0765E-1} & 8.3284E-1 & 4.3693E+0 &       & 10    & 1.8064E+1 & \cellcolor[rgb]{ .651,  .651,  .651}\textbf{1.6778E+1} & 1.8437E+1 & 2.7649E+2 \\
\cline{2-6}\cline{8-12}          & 30    & \cellcolor[rgb]{ .651,  .651,  .651}\textbf{1.4450E+0} & 1.8732E+0 & 2.7342E+0 & 1.0298E+1 &       & 30    & 1.2526E+2 & \cellcolor[rgb]{ .651,  .651,  .651}\textbf{9.1694E+1} & 1.6699E+2 & 5.3056E+3 \\
    \hline
    \multirow{1}[6]{*}{F2} & 2     & 7.7335E-1 & 8.1710E-1 & \cellcolor[rgb]{ .651,  .651,  .651}\textbf{7.6167E-1} & 1.9532E+0 & \multirow{1}[6]{*}{F12} & 2     & \cellcolor[rgb]{ .651,  .651,  .651}\textbf{1.2890E+6} & 1.3150E+6 & 2.1687E+6 & 2.3287E+7 \\
\cline{2-6}\cline{8-12}          & 10    & \cellcolor[rgb]{ .651,  .651,  .651}\textbf{1.7967E+2} & 1.9940E+2 & 2.4552E+2 & 3.6687E+3 &       & 10    & \cellcolor[rgb]{ .651,  .651,  .651}\textbf{1.2284E+7} & 1.2908E+7 & 1.7883E+7 & 4.9220E+8 \\
\cline{2-6}\cline{8-12}          & 30    & 1.8648E+3 & \cellcolor[rgb]{ .651,  .651,  .651}\textbf{1.5890E+3} & 2.9031E+3 & 2.4368E+5 &       & 30    & 1.7468E+8 & \cellcolor[rgb]{ .651,  .651,  .651}\textbf{1.7299E+8} & 2.3724E+8 & 7.7375E+9 \\
    \hline
    \multirow{1}[6]{*}{F3} & 2     & 2.5151E+1 & 2.6593E+1 & \cellcolor[rgb]{ .651,  .651,  .651}\textbf{1.7417E+1} & 3.0550E+1 & \multirow{1}[6]{*}{F13} & 2     & \cellcolor[rgb]{ .651,  .651,  .651}\textbf{4.0126E-1} & 6.3972E-1 & 8.5819E-1 & 3.8741E+0 \\
\cline{2-6}\cline{8-12}          & 10    & \cellcolor[rgb]{ .651,  .651,  .651}\textbf{8.5900E+2} & 1.1733E+3 & 9.8913E+2 & 1.5110E+4 &       & 10    & \cellcolor[rgb]{ .651,  .651,  .651}\textbf{6.4477E+0} & 1.8491E+1 & 7.6669E+0 & 1.6785E+2 \\
\cline{2-6}\cline{8-12}          & 30    & 8.8047E+3 & \cellcolor[rgb]{ .651,  .651,  .651}\textbf{7.7415E+3} & 1.2159E+4 & 7.9610E+5 &       & 30    & 4.5265E+1 & 6.9909E+1 & \cellcolor[rgb]{ .651,  .651,  .651}\textbf{2.6507E+1} & 7.9910E+2 \\
    \hline
    \multirow{1}[6]{*}{F4} & 2     & 2.2263E-1 & 2.5603E-1 & \cellcolor[rgb]{ .651,  .651,  .651}\textbf{2.1234E-1} & 4.1061E-1 & \multirow{1}[6]{*}{F14} & 2     & \cellcolor[rgb]{ .651,  .651,  .651}\textbf{3.4411E-1} & 3.5736E-1 & 3.9440E-1 & 4.1817E-1 \\
\cline{2-6}\cline{8-12}          & 10    & \cellcolor[rgb]{ .651,  .651,  .651}\textbf{2.4946E-1} & 2.6379E-1 & 3.5331E-1 & 1.1068E+0 &       & 10    & 2.3350E-2 & 7.1733E-2 & \cellcolor[rgb]{ .651,  .651,  .651}\textbf{2.0398E-2} & 4.7365E-2 \\
\cline{2-6}\cline{8-12}          & 30    & \cellcolor[rgb]{ .651,  .651,  .651}\textbf{7.6201E-2} & 1.5856E-1 & 1.4125E-1 & 6.4510E-1 &       & 30    & 1.6935E-2 & 7.1612E-2 & \cellcolor[rgb]{ .651,  .651,  .651}\textbf{1.4585E-2} & 6.7035E-2 \\
    \hline
    \multirow{1}[6]{*}{F5} & 2     & 1.0446E-2 & 1.1772E-2 & \cellcolor[rgb]{ .651,  .651,  .651}\textbf{1.0122E-2} & 1.1300E-2 & \multirow{1}[6]{*}{F15} & 2     & \cellcolor[rgb]{ .651,  .651,  .651}\textbf{2.3576E-1} & 2.4196E-1 & 2.6502E-1 & 3.1265E-1 \\
\cline{2-6}\cline{8-12}          & 10    & 2.7905E-2 & 2.7986E-2 & \cellcolor[rgb]{ .651,  .651,  .651}\textbf{2.7347E-2} & 9.4473E-2 &       & 10    & \cellcolor[rgb]{ .651,  .651,  .651}\textbf{1.0227E+0} & 1.0634E+0 & 1.2373E+0 & 3.0674E+0 \\
\cline{2-6}\cline{8-12}          & 30    & \cellcolor[rgb]{ .651,  .651,  .651}\textbf{3.3163E-2} & 3.6446E-2 & 4.8344E-2 & 2.5692E-1 &       & 30    & \cellcolor[rgb]{ .651,  .651,  .651}\textbf{3.2431E+0} & 4.1342E+0 & 5.0520E+0 & 3.4736E+1 \\
    \hline
    \multirow{1}[6]{*}{F6} & 2     & 2.1599E-1 & 2.3193E-1 & \cellcolor[rgb]{ .651,  .651,  .651}\textbf{2.0715E-1} & 3.0823E-1 & \multirow{1}[6]{*}{F16} & 2     & \cellcolor[rgb]{ .651,  .651,  .651}\textbf{2.8404E-1} & 3.1241E-1 & 3.2114E-1 & 3.9340E-1 \\
\cline{2-6}\cline{8-12}          & 10    & 1.4138E+0 & \cellcolor[rgb]{ .651,  .651,  .651}\textbf{1.3799E+0} & 1.5950E+0 & 4.9727E+0 &       & 10    & 1.3076E+0 & \cellcolor[rgb]{ .651,  .651,  .651}\textbf{1.2970E+0} & 1.5427E+0 & 4.6604E+0 \\
\cline{2-6}\cline{8-12}          & 30    & \cellcolor[rgb]{ .651,  .651,  .651}\textbf{2.6476E+0} & 2.8931E+0 & 3.7200E+0 & 1.5060E+1 &       & 30    & \cellcolor[rgb]{ .651,  .651,  .651}\textbf{4.3627E+0} & 5.3324E+0 & 6.6650E+0 & 6.0207E+1 \\
    \hline
    \multirow{1}[6]{*}{F7} & 2     & 5.6405E+0 & \cellcolor[rgb]{ .651,  .651,  .651}\textbf{5.5008E+0} & 7.5562E+0 & 6.5047E+1 & \multirow{1}[6]{*}{F17} & 2     & 6.5516E-2 & 7.2802E-2 & \cellcolor[rgb]{ .651,  .651,  .651}\textbf{5.8908E-2} & 6.8641E-2 \\
\cline{2-6}\cline{8-12}          & 10    & \cellcolor[rgb]{ .651,  .651,  .651}\textbf{6.8407E+1} & 6.8467E+1 & 8.7470E+1 & 1.8612E+3 &       & 10    & \cellcolor[rgb]{ .651,  .651,  .651}\textbf{1.8298E-1} & 1.9186E-1 & 1.9132E-1 & 2.3036E-1 \\
\cline{2-6}\cline{8-12}          & 30    & 5.1245E+2 & \cellcolor[rgb]{ .651,  .651,  .651}\textbf{4.4561E+2} & 5.4400E+2 & 1.7949E+4 &       & 30    & \cellcolor[rgb]{ .651,  .651,  .651}\textbf{4.1992E-1} & 5.7923E-1 & 4.9740E-1 & 1.8528E+0 \\
    \hline
    \multirow{1}[6]{*}{F8} & 2     & \cellcolor[rgb]{ .651,  .651,  .651}\textbf{6.4421E+0} & 8.2598E+0 & 1.0246E+1 & 7.5015E+1 & \multirow{1}[6]{*}{F18} & 2     & \cellcolor[rgb]{ .651,  .651,  .651}\textbf{7.9171E+0} & 8.8528E+0 & 9.5477E+0 & 5.0562E+1 \\
\cline{2-6}\cline{8-12}          & 10    & 7.3016E+1 & \cellcolor[rgb]{ .651,  .651,  .651}\textbf{6.8773E+1} & 9.2672E+1 & 2.8420E+3 &       & 10    & \cellcolor[rgb]{ .651,  .651,  .651}\textbf{3.8751E+2} & 3.9933E+2 & 6.9504E+2 & 1.9780E+4 \\
\cline{2-6}\cline{8-12}          & 30    & 5.8388E+2 & \cellcolor[rgb]{ .651,  .651,  .651}\textbf{4.7055E+2} & 7.4258E+2 & 2.1170E+4 &       & 30    & 6.8663E+3 & \cellcolor[rgb]{ .651,  .651,  .651}\textbf{5.9223E+3} & 9.8100E+3 & 3.7050E+5 \\
    \hline
    \multirow{1}[6]{*}{F9} & 2     & \cellcolor[rgb]{ .651,  .651,  .651}\textbf{1.2270E+6}       & 1.4330E+6 & 1.5548E+6 & 2.0123E+7      & \multirow{1}[6]{*}{F19} & 2     & 2.9912E-1 & 2.8090E-1 & \cellcolor[rgb]{ .651,  .651,  .651}\textbf{2.6204E-1} & 8.0894E-1 \\
\cline{2-6}\cline{8-12}    & 10    & \cellcolor[rgb]{ .651,  .651,  .651}\textbf{4.1771E+5} & 4.7615E+5 & 7.5221E+5 & 8.1209E+6      &       & 10    & \cellcolor[rgb]{ .651,  .651,  .651}\textbf{7.5440E-1} & 7.6484E-1 & 8.2718E-1 & 2.5802E+0 \\
\cline{2-6}\cline{8-12}    & 30    & 3.9609E+6 & \cellcolor[rgb]{ .651,  .651,  .651}\textbf{3.8603E+6} & 5.1305E+6 & 1.6692E+8 &       & 30    & 1.3788E+1 & \cellcolor[rgb]{ .651,  .651,  .651}\textbf{1.2999E+1} & 1.5938E+1 & 2.7295E+2 \\
    \hline
    \multirow{1}[6]{*}{F10} & 2     & 9.5085E+0 & \cellcolor[rgb]{ .651,  .651,  .651}\textbf{6.9783E+0} & 1.2881E+1 & 6.7627E+1 & \multirow{1}[6]{*}{F20} & 2     & 3.5455E-2 & 3.9468E-2 & \cellcolor[rgb]{ .651,  .651,  .651}\textbf{3.3128E-2} & 3.7422E-2 \\
\cline{2-6}\cline{8-12}          & 10    & 7.8455E+1 & \cellcolor[rgb]{ .651,  .651,  .651}\textbf{7.6291E+1} & 8.3358E+1 & 2.9098E+3 &       & 10    & 4.0848E-2 & 4.3134E-2 & \cellcolor[rgb]{ .651,  .651,  .651}\textbf{3.3434E-2} & 5.6918E-2 \\
\cline{2-6}\cline{8-12}          & 30    & 5.7160E+2 & \cellcolor[rgb]{ .651,  .651,  .651}\textbf{4.3714E+2} & 7.0516E+2 & 2.0198E+4 &       & 30    & \cellcolor[rgb]{ .651,  .651,  .651}\textbf{5.6556E-2} & 7.5748E-2 & 5.7657E-2 & 1.8357E-1 \\
    \hline
    \end{tabular}
}
\label{tab:emd}
\end{table*}


\section{Conclusions and Future Directions}
\label{sec:conclusion}

It is not uncommon a meta-heuristic algorithm is accompanied by some parameters, the settings of which largely influence its performance on various problems. Tweaking the parameter configuration of a meta-heuristic algorithm to achieve its peak performance on a certain problem can be treated as an optimisation process, as known as PO. Due to the stochastic property of most meta-heuristic algorithms, evaluating the quality of a particular parameter configuration usually requires to run the target algorithms several times. Therefore, it is inarguably that PO is computationally expensive. Building a cheap-to-evaluate surrogate model in lieu of a computationally expensive experiment has been widely accepted as a major approach for expensive optimisation. Instead of developing a new algorithm for PO, this paper aims to study a fundamental issue --- investigating the ability of four prevalent regression algorithms for building a surrogate model of empirical performance. From our extensive experiments, we find that surrogate models built by GP and RF have shown promising generalisation ability for predicting the empirical performance of unseen parameter configurations. In particular, the prediction accuracy depends on the quality of the original performance data. This implies that it needs to be careful to use a surrogate model in the early stage of a PO process. Furthermore, we find that although SVR does not show a promising performance for predicting the approximation error of a parameter configuration, it is able to differentiate the order of two parameter configurations. 

Generally speaking, we hope this work will be useful to a wide variety of researchers who seek to model algorithm performance for algorithm analysis, scheduling, algorithm portfolio construction, automated algorithm configuration, and other applications. As for the coming next step, we plan to explore the following three aspects.
\vspace{-0.2em} 
\begin{itemize}
    \item We would like to apply the regression algorithms investigated in this paper in the context of model-based PO. Although using design and analysis of computer experiments in the context of PO has already been studied in some previous work (e.g. sequential PO~\cite{Bartz-BeielsteinLP05}), it is still worthwhile to see whether the observations in the offline training are directly applicable to online PO.
    \item Since collecting a performance data in PO is computationally expensive, it might be interesting to use the offline trained surrogate models to generate pseudo data. In this rigour, semi-supervised learning~\cite{SunGJC13} can be useful to address a small data challenge.
    \item Here we set the PO as a per-instance scenario. In the prevalent algorithm configuration literature~\cite{HutterHL11}, it is more interesting to combine the problem feature into the surrogate modelling process so that we can generalise the PO to a range of similar problems.
    \item In addition, assessing the performance of evolutionary multi-objective optimisation algorithms, e.g.~\cite{WuKZLWL15,WuLKZZ17,WuLKZ18,LiCFY18,LiDY18}, is even more difficult. Therefore, it is also interesting to investigate appropriate surrogate modelling methods to analyse and understanding the parameter versus algorithm performance in the context of multi-objective optimisation.
\end{itemize}
\vspace{-0.5em}


\section*{Acknowledgment}
This work was supported by the Royal Society (Grant No. IEC/NSFC/170243).

\bibliographystyle{IEEEtran}
\bibliography{IEEEabrv,epm}

\begin{thebibliography}{10}
\providecommand{\url}[1]{#1}
\csname url@samestyle\endcsname
\providecommand{\newblock}{\relax}
\providecommand{\bibinfo}[2]{#2}
\providecommand{\BIBentrySTDinterwordspacing}{\spaceskip=0pt\relax}
\providecommand{\BIBentryALTinterwordstretchfactor}{4}
\providecommand{\BIBentryALTinterwordspacing}{\spaceskip=\fontdimen2\font plus
\BIBentryALTinterwordstretchfactor\fontdimen3\font minus
  \fontdimen4\font\relax}
\providecommand{\BIBforeignlanguage}[2]{{%
\expandafter\ifx\csname l@#1\endcsname\relax
\typeout{** WARNING: IEEEtran.bst: No hyphenation pattern has been}%
\typeout{** loaded for the language `#1'. Using the pattern for}%
\typeout{** the default language instead.}%
\else
\language=\csname l@#1\endcsname
\fi
#2}}
\providecommand{\BIBdecl}{\relax}
\BIBdecl

\bibitem{Jin11}
Y.~Jin, ``Surrogate-assisted evolutionary computation: Recent advances and
  future challenges,'' \emph{Swarm and Evol. Comput.}, vol.~1, no.~2, pp.
  61--70, 2011.

\bibitem{SantnerWN03}
T.~J. Santner, B.~J. Williams, and W.~I. Notz, \emph{The Design and Analysis of
  Computer Experiments}.\hskip 1em plus 0.5em minus 0.4em\relax Springer, 2003.

\bibitem{Bartz-BeielsteinLP05}
T.~Bartz{-}Beielstein, C.~Lasarczyk, and M.~Preuss, ``Sequential parameter
  optimization,'' in \emph{CEC'05: Proc. of the 2005 {IEEE} Congress on Evol.
  Comput.}, 2005, pp. 773--780.

\bibitem{HutterHL11}
F.~Hutter, H.~H. Hoos, and K.~Leyton{-}Brown, ``Sequential model-based
  optimization for general algorithm configuration,'' in \emph{LION'11: Proc.
  of 5th International Conference on Learning and Intelligent Optimization},
  2011, pp. 507--523.

\bibitem{ThorntonHHL13}
C.~Thornton, F.~Hutter, H.~H. Hoos, and K.~Leyton{-}Brown, ``Auto-{WEKA}:
  combined selection and hyperparameter optimization of classification
  algorithms,'' in \emph{KDD'13: Proc. of 19th {ACM} {SIGKDD} International
  Conference on Knowledge Discovery and Data Mining}, 2013, pp. 847--855.

\bibitem{BergstraB12}
J.~Bergstra and Y.~Bengio, ``Random search for hyper-parameter optimization,''
  \emph{J. Machine Learning Research}, vol.~13, pp. 281--305, 2012.

\bibitem{Reif14}
M.~Reif, F.~Shafait, M.~Goldstein, T.~Breuel, and A.~Dengel, ``Automatic
  classifier selection for non-experts,'' \emph{Pattern Anal. Appl.}, vol.~17,
  no.~1, pp. 83--96, 2014.

\bibitem{KohaviJ95}
R.~Kohavi and G.~H. John, ``Autmatic parameter selection by minimizing
  estimated error,'' in \emph{ICML'95: Proc. of 12th International Conference
  on Machine Learning}, 1995, pp. 304--312.

\bibitem{BlotHJKT16}
A.~Blot, H.~H. Hoos, L.~Jourdan, M.~Kessaci{-}Marmion, and H.~Trautmann,
  ``Mo-paramils: {A} multi-objective automatic algorithm configuration
  framework,'' in \emph{LION'16: Proc. of 10th International Conference on
  Learning and Intelligent Optimization}, 2016, pp. 32--47.

\bibitem{LopDubPerStuBir2016irace}
M.~L{\'o}pez-Ib{\'a}{\~n}ez, J.~Dubois-Lacoste, L.~{P{\'e}rez C{\'a}ceres},
  T.~St{\"u}tzle, and M.~Birattari, ``The irace package: Iterated racing for
  automatic algorithm configuration,'' \emph{Oper. Res Perspectives}, vol.~3,
  pp. 43--58, 2016.

\bibitem{LiFKZ14}
K.~Li, {\'{A}}.~Fialho, S.~Kwong, and Q.~Zhang, ``Adaptive operator selection
  with bandits for a multiobjective evolutionary algorithm based on
  decomposition,'' \emph{{IEEE} Trans. Evolutionary Computation}, vol.~18,
  no.~1, pp. 114--130, 2014.

\bibitem{SnoekLA12}
J.~Snoek, H.~Larochelle, and R.~P. Adams, ``Practical bayesian optimization of
  machine learning algorithms,'' in \emph{NIPS'12: Proc. of 26th Annual
  Conference on Neural Information Processing Systems}, 2012, pp. 2960--2968.

\bibitem{SandersG17}
S.~Sanders and C.~G. Giraud{-}Carrier, ``Informing the use of hyperparameter
  optimization through metalearning,'' in \emph{ICDM'17: Proc. of 2017 {IEEE}
  International Conference on Data Mining}, 2017, pp. 1051--1056.

\bibitem{CaoKWL12}
J.~Cao, S.~Kwong, R.~Wang, and K.~Li, ``A weighted voting method using minimum
  square error based on extreme learning machine,'' in \emph{ICMLC'12: Proc. of
  the 2012 International Conference on Machine Learning and Cybernetics}, 2012,
  pp. 411--414.

\bibitem{LiWKC13}
K.~Li, R.~Wang, S.~Kwong, and J.~Cao, ``Evolving extreme learning machine
  paradigm with adaptive operator selection and parameter control,''
  \emph{International Journal of Uncertainty, Fuzziness and Knowledge-Based
  Systems}, vol.~21, pp. 143--154, 2013.

\bibitem{CaoKWL14}
J.~Cao, S.~Kwong, R.~Wang, and K.~Li, ``{AN} indicator-based selection
  multi-objective evolutionary algorithm with preference for multi-class
  ensemble,'' in \emph{ICMLC'14: Proc. of the 2014 International Conference on
  Machine Learning and Cybernetics}, 2014, pp. 147--152.

\bibitem{CaoKWLLK15}
J.~Cao, S.~Kwong, R.~Wang, X.~Li, K.~Li, and X.~Kong, ``Class-specific soft
  voting based multiple extreme learning machines ensemble,''
  \emph{Neurocomputing}, vol. 149, pp. 275--284, 2015.

\bibitem{automl}
``Neurips 2018 challenge: The 3rd automl challenge: Automl for lifelong machine
  learning,'' \url{https://www.4paradigm.com/competition/nips2018}.

\bibitem{StornP97}
R.~Storn and K.~V. Price, ``Differential evolution - {A} simple and efficient
  heuristic for global optimization over continuous spaces,'' \emph{J. Global
  Optimization}, vol.~11, no.~4, pp. 341--359, 1997.

\bibitem{LiKWCR12}
K.~Li, S.~Kwong, R.~Wang, J.~Cao, and I.~J. Rudas, ``Multi-objective
  differential evolution with self-navigation,'' in \emph{SMC'12: Proc. of the
  2012 {IEEE} International Conference on Systems, Man, and Cybernetics}, 2012,
  pp. 508--513.

\bibitem{SuganthanHDLCAT05}
P.~N. Suganthan, N.~Hansen, K.~Deb, J.~J. Liang, Y.-P. Chen, A.~Anger, and
  S.~Tiwari, ``Problem definitions and evaluation criteria for the {CEC} 2005
  special session on real-parameter optimization,'' NTU and IIT Kanpur,
  Technical Report 2005005, 2005.

\bibitem{LiZKLW14}
K.~Li, Q.~Zhang, S.~Kwong, M.~Li, and R.~Wang, ``Stable matching-based
  selection in evolutionary multiobjective optimization,'' \emph{{IEEE} Trans.
  Evolutionary Computation}, vol.~18, no.~6, pp. 909--923, 2014.

\bibitem{LiKD15}
K.~Li, S.~Kwong, and K.~Deb, ``A dual-population paradigm for evolutionary
  multiobjective optimization,'' \emph{Inf. Sci.}, vol. 309, pp. 50--72, 2015.

\bibitem{LiKZD15}
K.~Li, S.~Kwong, Q.~Zhang, and K.~Deb, ``Interrelationship-based selection for
  decomposition multiobjective optimization,'' \emph{{IEEE} Trans.
  Cybernetics}, vol.~45, no.~10, pp. 2076--2088, 2015.

\bibitem{LiDZK15}
K.~Li, K.~Deb, Q.~Zhang, and S.~Kwong, ``An evolutionary many-objective
  optimization algorithm based on dominance and decomposition,'' \emph{{IEEE}
  Trans. Evolutionary Computation}, vol.~19, no.~5, pp. 694--716, 2015.

\bibitem{LiDZZ17}
K.~Li, K.~Deb, Q.~Zhang, and Q.~Zhang, ``Efficient nondomination level update
  method for steady-state evolutionary multiobjective optimization,''
  \emph{{IEEE} Trans. Cybernetics}, vol.~47, no.~9, pp. 2838--2849, 2017.

\bibitem{ChenLY18}
R.~Chen, K.~Li, and X.~Yao, ``Dynamic multiobjectives optimization with a
  changing number of objectives,'' \emph{{IEEE} Trans. Evolutionary
  Computation}, vol.~22, no.~1, pp. 157--171, 2018.

\bibitem{LiCMY18}
K.~Li, R.~Chen, G.~Min, and X.~Yao, ``Integration of preferences in
  decomposition multiobjective optimization,'' \emph{{IEEE} Trans.
  Cybernetics}, vol.~48, no.~12, pp. 3359--3370, 2018.

\bibitem{DasS11}
S.~Das and P.~N. Suganthan, ``Differential evolution: {A} survey of the
  state-of-the-art,'' \emph{{IEEE} Trans. Evol. Comput.}, vol.~15, no.~1, pp.
  4--31, 2011.

\bibitem{BrestGBMZ06}
J.~Brest, S.~Greiner, B.~Boskovic, M.~Mernik, and V.~Zumer, ``Self-adapting
  control parameters in differential evolution: {A} comparative study on
  numerical benchmark problems,'' \emph{{IEEE} Trans. Evol. Comput.}, vol.~10,
  no.~6, pp. 646--657, 2006.

\bibitem{QinHS09}
A.~K. Qin, V.~L. Huang, and P.~N. Suganthan, ``Differential evolution algorithm
  with strategy adaptation for global numerical optimization,'' \emph{{IEEE}
  Trans. Evol. Comput.}, vol.~13, no.~2, pp. 398--417, 2009.

\bibitem{LiFK11}
K.~Li, {\'{A}}.~Fialho, and S.~Kwong, ``Multi-objective differential evolution
  with adaptive control of parameters and operators,'' in \emph{LION'11: Proc.
  of 5th International Conference on Learning and Intelligent Optimization},
  2011, pp. 473--487.

\bibitem{BelkhirDSS16}
N.~Belkhir, J.~Dr{\'{e}}o, P.~Sav{\'{e}}ant, and M.~Schoenauer, ``Feature based
  algorithm configuration: {A} case study with differential evolution,'' in
  \emph{PPSN'16: Proc. of 14th International Conference on Parallel Problem
  Solving from Nature - {PPSN} {XIV}}, 2016, pp. 156--166.

\bibitem{HutterXHL14}
F.~Hutter, L.~Xu, H.~H. Hoos, and K.~Leyton{-}Brown, ``Algorithm runtime
  prediction: Methods {\&} evaluation,'' \emph{Artif. Intell.}, vol. 206, pp.
  79--111, 2014.

\bibitem{WuKJLZ17}
M.~Wu, S.~Kwong, Y.~Jia, K.~Li, and Q.~Zhang, ``Adaptive weights generation for
  decomposition-based multi-objective optimization using gaussian process
  regression,'' in \emph{Proceedings of the Genetic and Evolutionary
  Computation Conference, {GECCO} 2017, Berlin, Germany, July 15-19, 2017},
  2017, pp. 641--648.

\bibitem{WuLKZZ18}
M.~Wu, K.~Li, S.~Kwong, Q.~Zhang, and Q.~Zhang, ``Learning to decompose: a
  paradigm for decomposition-based multiobjective optimization,'' \emph{{IEEE}
  Trans. Evolutionary Computation}, 2018, accepted for publication.

\bibitem{LoshchilovSS10}
I.~Loshchilov, M.~Schoenauer, and M.~Sebag, ``Comparison-based optimizers need
  comparison-based surrogates,'' in \emph{PPSN'10: Proc. of 11th International
  Conference on Parallel Problem Solving from Nature}, 2010, pp. 364--373.

\bibitem{RubnerTG00}
Y.~Rubner, C.~Tomasi, and L.~J. Guibas, ``The earth mover's distance as a
  metric for image retrieval,'' \emph{International Journal of Computer
  Vision}, vol.~40, no.~2, pp. 99--121, 2000.

\bibitem{SunGJC13}
X.~Sun, D.~Gong, Y.~Jin, and S.~Chen, ``A new surrogate-assisted interactive
  genetic algorithm with weighted semisupervised learning,'' \emph{{IEEE}
  Trans. Cybernetics}, vol.~43, no.~2, pp. 685--698, 2013.

\bibitem{WuKZLWL15}
M.~Wu, S.~Kwong, Q.~Zhang, K.~Li, R.~Wang, and B.~Liu, ``Two-level stable
  matching-based selection in {MOEA/D},'' in \emph{SMC'15: Proc. of the 2015
  {IEEE} International Conference on Systems, Man, and Cybernetics}, 2015, pp.
  1720--1725.

\bibitem{WuLKZZ17}
M.~Wu, K.~Li, S.~Kwong, Y.~Zhou, and Q.~Zhang, ``Matching-based selection with
  incomplete lists for decomposition multiobjective optimization,''
  \emph{{IEEE} Trans. Evolutionary Computation}, vol.~21, no.~4, pp. 554--568,
  2017.

\bibitem{WuLKZ18}
M.~Wu, K.~Li, S.~Kwong, and Q.~Zhang, ``Evolutionary many-objective
  optimization based on adversarial decomposition,'' \emph{{IEEE} Trans.
  Cybernetics}, 2018, accepted for publication.

\bibitem{LiCFY18}
K.~Li, R.~Chen, G.~Fu, and X.~Yao, ``Two-archive evolutionary algorithm for
  constrained multi-objective optimization,'' \emph{{IEEE} Trans. Evolutionary
  Computation}, 2018, accepted for publication.

\bibitem{LiDY18}
K.~Li, K.~Deb, and X.~Yao, ``R-metric: Evaluating the performance of
  preference-based evolutionary multiobjective optimization using reference
  points,'' \emph{{IEEE} Trans. Evolutionary Computation}, vol.~22, no.~6, pp.
  821--835, 2018.

\end{thebibliography}

\end{document}